\documentclass[sigconf]{acmart}

\AtBeginDocument{%
  }

\usepackage{bbding}
\usepackage{cleveref}
\usepackage{multirow}
\usepackage{booktabs}    
\usepackage{multirow}    
\usepackage{graphicx}    
\usepackage{subcaption}  
\usepackage{xspace}
\usepackage{pifont}      
\usepackage{xcolor}      

\usepackage{amssymb}
\usepackage{pifont}

\usepackage{enumitem}
\definecolor{cmark}{RGB}{34,139,34}

\usepackage[table,xcdraw]{xcolor}
\usepackage{multirow}
\usepackage{tabularx}
\usepackage{float}
\usepackage{graphicx}
\usepackage{amsmath,mathrsfs}
\usepackage{amssymb}
\hypersetup{colorlinks=true, allcolors=red}

\newcommand{\eg}{\textit{e.g.}\xspace}
\newcommand{\ie}{\textit{i.e.}\xspace}

\definecolor{best}{RGB}{197,17,17}
\definecolor{second}{RGB}{0,82,180}
\newcommand{\first}[1]{\textbf{\textcolor{best}{#1}}}
\newcommand{\Second}[1]{\textcolor{second}{#1}}

\copyrightyear{2026}
\acmYear{2026}
\setcopyright{cc}
\setcctype{by-nc-nd}

\acmConference[MM '26]
  {Proceedings of the 34th ACM International Conference on Multimedia}
  {November 10--14, 2026}
  {Rio de Janeiro, Brazil}

\acmBooktitle{Proceedings of the 34th ACM International Conference on Multimedia
  (MM '26), November 10--14, 2026, Rio de Janeiro, Brazil}

\acmDOI{10.1145/3767308.3836546}
\acmISBN{979-8-4007-2213-4/2026/11}

\begin{document}

\title{HiCo-GS: Hierarchical Context Aggregation and Geometric Consistency for Octree Gaussian Splatting}


\author{Wei Zhang}
\authornote{These authors contributed equally to this work.}
\email{zhangwei707@mail.nwpu.edu.cn}
\affiliation{%
  \department{School of Computer Science}
  \institution{Northwestern Polytechnical University}
  \city{Xi'an}
  \country{China}
}

\author{Shengkai Yu}
\authornotemark[1]
\email{yyusober@mail.nwpu.edu.cn}
\affiliation{%
  \institution{Northwestern Polytechnical University}
  \city{Xi'an}
  \country{China}
}

\author{Shiqiang Gong}
\authornotemark[1]
\email{gongshiqiang@mail.nwpu.edu.cn}
\affiliation{%
  \institution{Northwestern Polytechnical University}
  \city{Xi'an}
  \country{China}
}

\author{Qi Zhang}
\email{nwpuqzhang@gmail.com}
\affiliation{%
  \department{vivo BlueImage Lab}
  \institution{vivo Mobile Communication Co., Ltd.}
  \city{Hangzhou}
  \country{China}
}

\author{Qiang Li}
\email{qiangli@nwpu.edu.cn}
\affiliation{%
  \institution{Northwestern Polytechnical University}
  \city{Xi'an}
  \country{China}
}

\author{Qi Wang}
\authornote{Corresponding author.}
\email{crabwq@gmail.com}
\affiliation{%
  \institution{Northwestern Polytechnical University}
  \city{Xi'an}
  \country{China}
}

\renewcommand{\shortauthors}{Zhang et al.}

\begin{abstract}
   Octree-based anchor Gaussian Splatting has emerged as a scalable representation for city-scale novel view synthesis, where multi-level anchors adaptively capture scene content from coarse building structures to fine architectural details. However, we identify a fundamental limitation in existing methods: cross-level feature isolation, where each level's anchor features are optimized independently with no inter-level communication, causing color drift on building facades and over-smoothing in textured regions. We present HiCo-GS, a high-fidelity reconstruction framework with two complementary modules. Cross-Level Context Aggregation (CLCA) enables bidirectional hierarchical prior injection by leveraging the octree's spatial containment structure to aggregate per-level context vectors into parent-self-child triplets, fused via a lightweight MLP with residual connection. Coarse-level structural priors flow down to inform fine-level anchors, while fine-level detail statistics feed back to prevent over-smoothing, at negligible computational overhead. Depth-Normal Geometric Consistency (DNGC) regularization enforces agreement between rendered normals and depth-derived normals through an alpha-weighted consistency loss, complemented by edge-aware smoothness losses with progressive warmup that exploit the strong planar priors ubiquitous in urban geometry to suppress floating artifacts. We further introduce the China-Pagoda dataset comprising 8 ancient Chinese pagodas with over 1,200 images each, featuring dense ornamental carvings, curved multi-layer eaves, and repetitive fine-grained textures. Extensive experiments on Mill19, UrbanScene3D, MatrixCity, and China-Pagoda demonstrate that HiCo-GS achieves state-of-the-art rendering quality and substantially cleaner geometry across real-world and synthetic urban benchmarks.  \textbf{Code:}~{\small\url{https://github.com/WZ-CS/HiCo-GS}}.
\end{abstract}



\begin{teaserfigure}
  \centering
  \includegraphics[width=0.85\textwidth]{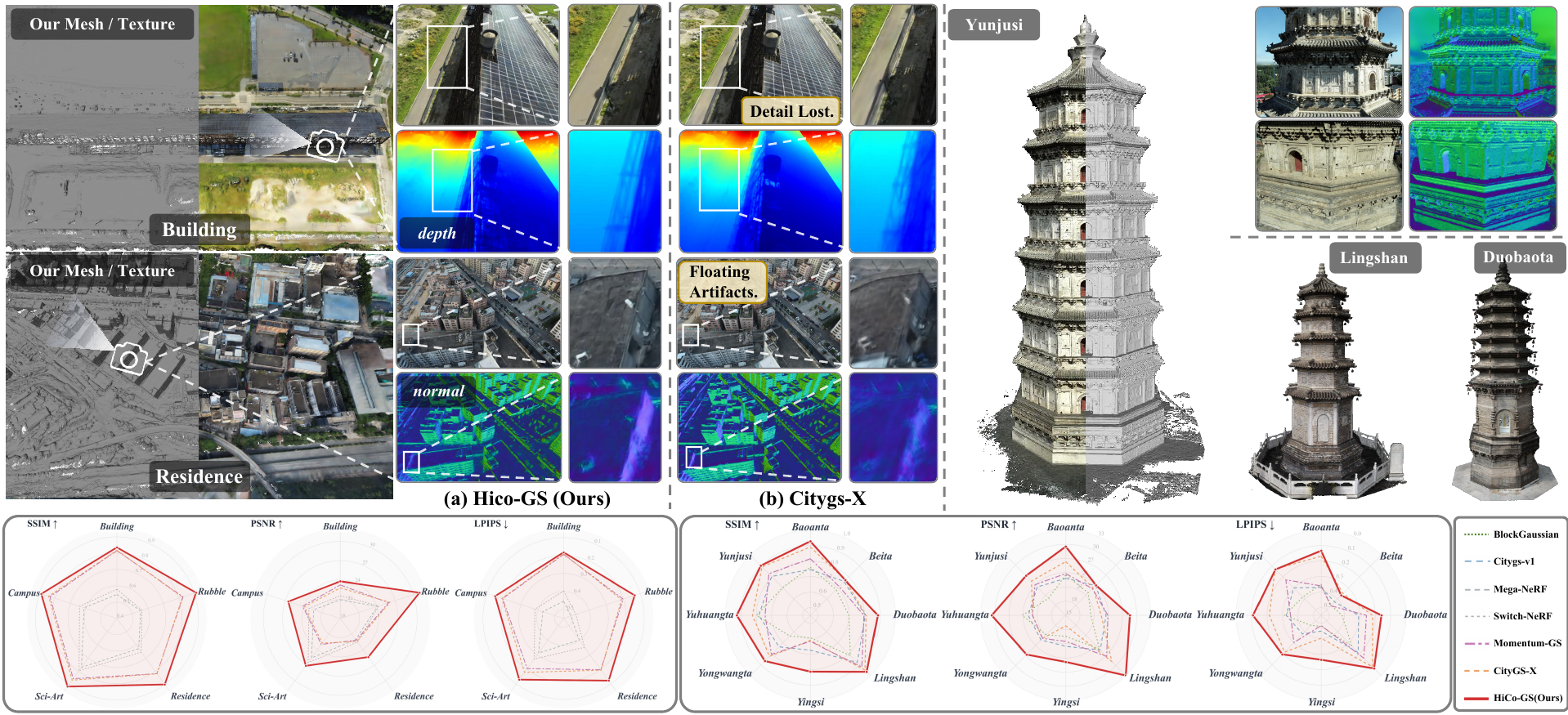}
  \caption{\textbf{HiCo-GS} achieves high-fidelity urban reconstruction with cleaner geometry. \textit{Left:} Mesh, rendered RGB, depth, and normal comparisons on Mill19 and UrbanScene3D between (a) HiCo-GS and (b) CityGS-$\mathcal{X}$. \textit{Right:} Representative scenes from our China-Pagoda dataset featuring extreme architectural complexity. \textit{Bottom:} Radar charts summarizing quantitative results across both benchmarks. HiCo-GS (red) consistently outperforms prior methods.}
  \label{fig:teaser}
\end{teaserfigure}


\maketitle

\vspace{-1.6 mm}


\begin{figure}[t]
    \centering
    \includegraphics[width=0.8\linewidth]{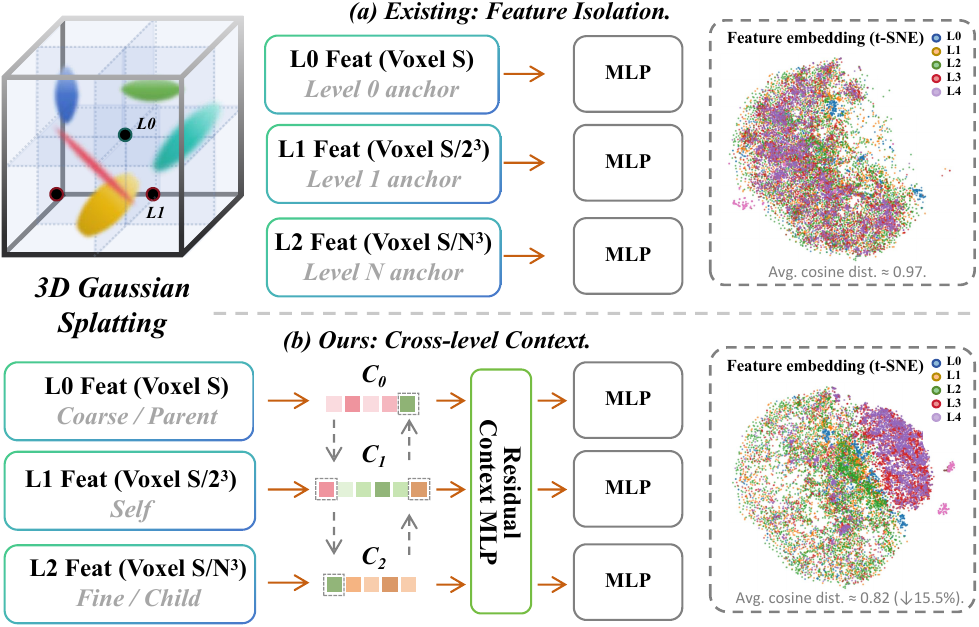}
    \caption{\textbf{Cross-level feature isolation and our solution.} \textit{Top:} In existing octree-based anchor Gaussian Splatting, multi-level anchors (L0 to LN) are optimized independently---each level's features are decoded by shared MLPs with no inter-level communication. The t-SNE visualization (\textit{right}) confirms this: features from all levels form a structureless mixture with inter-level cosine distances approaching 0.97, indicating near-orthogonal representations. \textit{Bottom:} Our CLCA enables bidirectional hierarchical prior injection. Each anchor receives a context triplet from its parent (coarse), self, and child (fine) levels, fused through a lightweight MLP with residual connection. Coarse-level structural priors flow down while fine-level detail statistics feed back. After CLCA (\textit{right}), features exhibit level-aware clustering with adjacent-level cosine distance reduced by up to 43\%.}
    \label{fig:motivation}
\end{figure}

\section{Introduction}
\label{sec:intro}

3D Gaussian Splatting (3DGS)~\cite{kerbl20233d,zhang2025review} has established a new paradigm for real-time novel view synthesis by representing scenes as collections of explicit Gaussian primitives rendered through tile-based differentiable rasterization. To scale this representation to city-level environments~\cite{zhaosigma,zhao2025rli}, recent works~\cite{lu2024scaffold,ren2024octree,liu2024citygaussian,gao2025citygs} adopt octree-based anchor structures with level-of-detail (LOD) control, where multi-level anchors reside at different voxel resolutions. Coarse-level anchors capture building-scale geometry, while fine-level anchors encode local architectural details such as window frames and surface textures. Neural Gaussian generation then decodes each anchor's learned feature into a local cluster of Gaussians conditioned on the viewing direction, enabling compact yet expressive scene representation. These methods have demonstrated strong scalability through distributed training strategies, making large-scale urban reconstruction increasingly practical.

Despite their success, we identify two complementary limitations in existing octree-anchor methods. The first is \textit{cross-level feature isolation}: as illustrated in Fig.~\ref{fig:motivation} (top), each level's anchor features are optimized independently through backpropagation with no inter-level communication. A fine-level anchor encoding a window frame has no access to the coarse-level context of the wall it belongs to, and conversely, a coarse-level anchor is unaware of the local detail density beneath it. This isolation leads to color drift on building facades and over-smoothing in richly textured regions. The second limitation is the \textit{absence of geometric supervision}: the rasterization stage produces two independent normal estimates, one from Gaussian covariance and one from the rendered depth map, yet existing methods impose no constraint requiring these signals to agree. In urban scenes dominated by large planar surfaces, this lack of geometric consistency results in noisy normals, depth discontinuities, and floating artifacts.

We present \textbf{HiCo-GS}, a high-fidelity reconstruction framework that addresses both limitations. To resolve cross-level feature isolation, we propose \textbf{Cross-Level Context Aggregation (CLCA)}, which leverages the octree's inherent spatial containment structure to enable bidirectional context flow across levels. For each anchor, CLCA looks up the specific parent-level context vector at the coarser voxel that spatially contains it, and aggregates child-level features from finer voxels that fall within its own grid cell. The resulting parent-self-child context triplet is fused through a lightweight MLP with residual connection, allowing coarse-level structural priors to inform fine-level anchors while fine-level detail statistics feed back to coarse-level anchors (Fig.~\ref{fig:motivation}, bottom). To address the lack of geometric supervision, we introduce \textbf{Depth-Normal Geometric Consistency (DNGC)} regularization, which enforces agreement between rendered normals and depth-derived normals through an alpha-weighted consistency loss, complemented by edge-aware smoothness terms that encourage planarity on smooth surfaces while preserving sharp boundaries. A progressive warmup schedule ensures that geometric constraints do not interfere with early photometric convergence. Furthermore, to stress-test reconstruction fidelity on geometrically extreme architectures beyond standard urban benchmarks, we introduce the \textbf{China-Pagoda} dataset comprising 8 ancient Chinese pagodas with over 1,200 images each, featuring dense ornamental carvings, curved multi-layer eaves, and repetitive fine-grained textures.
Our contributions are as follows:
\begin{itemize}
\item We identify the cross-level feature isolation problem in octree-based Gaussian Splatting and propose CLCA, a spatially aware context aggregation module that enables bidirectional hierarchical feature communication across octree levels by exploiting the octree's spatial containment structure, enriching fine-level features with structural priors and coarse-level features with detail statistics.
\item We introduce DNGC, a depth-normal geometric consistency regularization with edge-aware smoothness and progressive warmup that leverages urban planar priors to suppress floating artifacts and produce cleaner surface geometry.
\item We construct the China-Pagoda benchmark for evaluating reconstruction under extreme geometric complexity, and demonstrate through extensive experiments on Mill19, UrbanScene3D, MatrixCity, and China-Pagoda that HiCo-GS achieves state-of-the-art rendering quality and substantially cleaner geometry across real-world and synthetic urban benchmarks.
\end{itemize}

\begin{figure*}[t]
    \centering
    \includegraphics[width=0.80\linewidth]{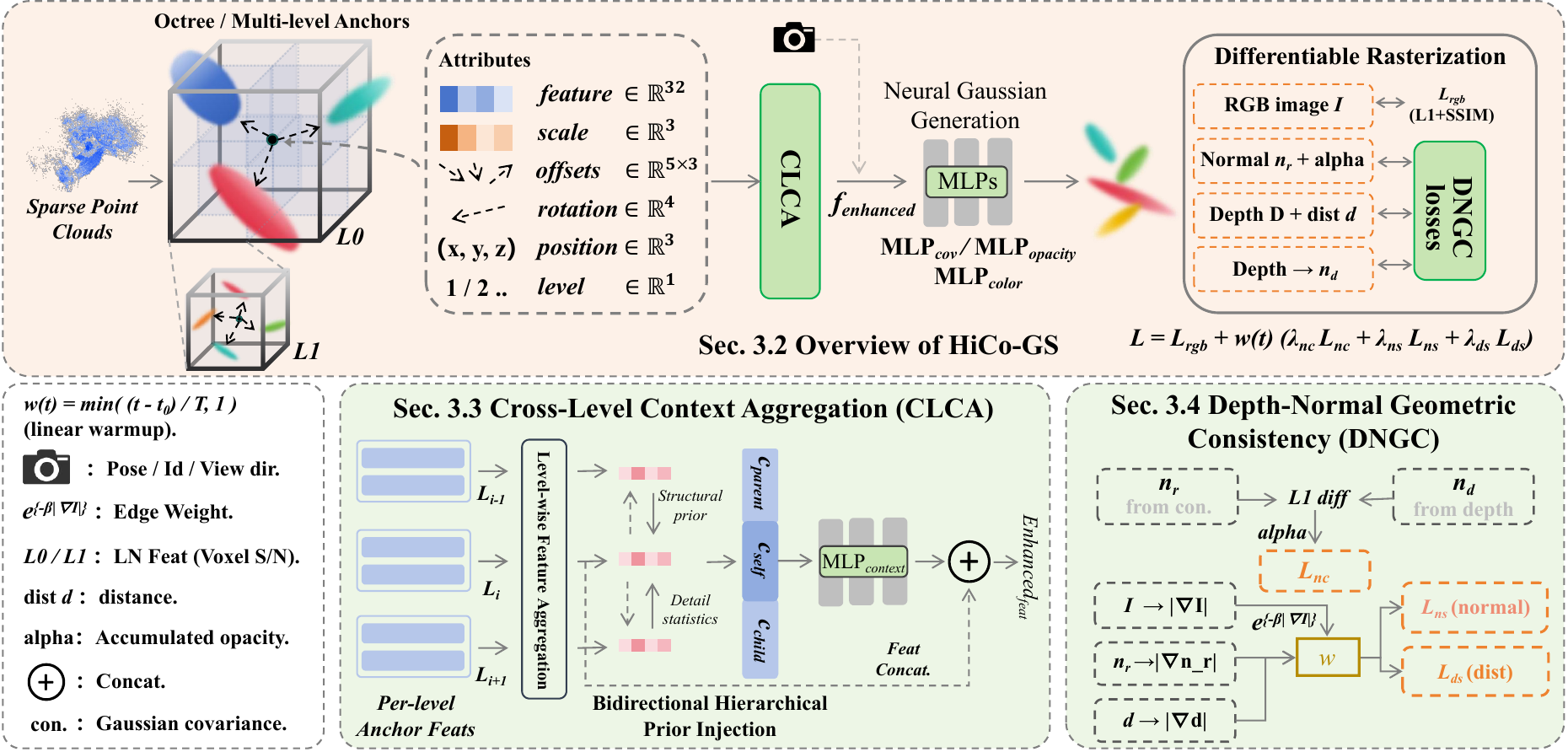}
    \caption{\textbf{Overview of HiCo-GS.}
    \textbf{(a)} A sparse point cloud is organized into a multi-level octree, where each anchor stores a feature vector, position, level, offsets, scale, and rotation. Visible anchors are enhanced by CLCA to produce $\hat{\mathbf{f}}$, decoded by three shared MLPs (conditioned on viewing direction and appearance) into neural Gaussians. Differentiable rasterization produces RGB image $\mathbf{I}$, rendered normal $\mathbf{n}_r$, depth $\mathbf{D}$ (with derived normal $\mathbf{n}_d$), distance $\mathbf{d}$, and opacity $\boldsymbol{\alpha}$. \textbf{(b)} CLCA performs spatially-aware level-wise aggregation to compute per-level context vectors, then assembles a parent-self-child triplet for each anchor via hierarchical context lookup. The triplet is fused by $\mathrm{MLP}_{\text{context}}$ with a residual connection. \textbf{(c)} DNGC enforces consistency between $\mathbf{n}_r$ (from Gaussian covariance) and $\mathbf{n}_d$ (from depth) via $\mathcal{L}_{\text{nc}}$, and applies edge-aware smoothness losses $\mathcal{L}_{\text{ns}}$, $\mathcal{L}_{\text{ds}}$ weighted by $e^{-\beta|\nabla\mathbf{I}|}$ to enforce planarity while preserving sharp boundaries.}
    \label{fig:pipeline}
\end{figure*}

\section{Related Work}
\label{sec:related}
\noindent\textbf{3D Gaussian Splatting and Large-Scale Extensions.}
3DGS~\cite{kerbl20233d} represents scenes as anisotropic Gaussian primitives rendered via differentiable rasterization, with subsequent works improving compactness~\cite{niedermayr2024compressed,elrawy2025opacity,fan2024lightgaussian,zhang2025refined, 11595022}, anti-aliasing~\cite{yan2024multi,yu2024mip}, and appearance modeling~\cite{dahmani2024swag,kulhanek2024wildgaussians,zhang2025semantic}. Scaffold-GS~\cite{lu2024scaffold} introduces anchor-based neural Gaussians, which our work builds upon. Scaling to city-level scenes has been addressed via spatial partitioning~\cite{lin2024vastgaussian,liu2024citygaussian,xu2024grid4d,chen2024dogs}, hierarchical LOD~\cite{kerbl2024hierarchical,ren2024octree,shen2025lod,zoomers2025progs,seo2024flod}, and distributed training~\cite{gao2025citygs}. Notably, Octree-GS~\cite{ren2024octree} organizes anchors into a multi-level octree with LOD-aware densification, and Hierarchy-GS~\cite{kerbl2024hierarchical} enables smooth transitions via chunk-based consolidation. 

\noindent\textbf{Geometric Regularization for Gaussian Splatting.}
Several works incorporate geometric priors to improve surface quality: SuGaR~\cite{guedon2024sugar} aligns Gaussians to surfaces for mesh extraction, GOF~\cite{yu2024gaussian} derives opacity fields for surface extraction, and normal-based supervision has been explored via rendered consistency~\cite{turkulainen2025dn,jiang2024gaussianshader,zhang2024visual} and mono-depth priors~\cite{chung2024depth}. For urban scenes, CityGaussianV2~\cite{liu2024citygaussianv2} and ULSR-GS~\cite{li2024ulsr} enforce geometric consistency across partitions. Our DNGC enforces consistency between rendered normals and depth-derived normals already produced by the rasterizer, using edge-aware weights that respect urban planar structure.

\section{Method}

\subsection{Preliminaries}
\label{sec:prelim}

\noindent\textbf{Anchor-based Octree Gaussian Splatting.}
Anchor-based methods~\cite{lu2024scaffold,ren2024octree} place anchors at octree voxel centers, each storing a learnable feature $\mathbf{f}_i \in \mathbb{R}^d$, position $\mathbf{p}_i$, scaling $\mathbf{s}_i \in \mathbb{R}^6$, rotation $\mathbf{q}_i \in \mathbb{R}^4$, opacity $o_i$, and $K$ offset vectors $\{\boldsymbol{\delta}_i^k\}_{k=1}^{K}$. Each anchor generates $K$ neural Gaussians centered at $\mathbf{x}_i^k = \mathbf{p}_i + \boldsymbol{\delta}_i^k \odot \mathbf{s}_i^{[1:3]}$, whose color, opacity, and covariance are predicted by shared MLPs conditioned on $\mathbf{f}_i$ and the viewing direction. 
Given a camera at position $\mathbf{c}$, each anchor's predicted level is
\begin{equation}
    \hat{l}_i = \left\lfloor \frac{\log_2(d_{\text{std}} / \|\mathbf{p}_i - \mathbf{c}\|)}{\log_2 b} + \epsilon_i \right\rceil,
    \label{eq:lod}
\end{equation}
where $d_{\text{std}}$ is a standard distance from training cameras and $\epsilon_i$ a learnable adjustment. An anchor at level $l_i$ is visible only if $l_i \leq \hat{l}_i$.

\noindent\textbf{Rasterization Outputs.}
Differentiable rasterization produces the RGB image $\mathbf{I}$, accumulated opacity $\boldsymbol{\alpha}$, and two independent normal estimates. The \textit{rendered normal} $\mathbf{n}_r$ is obtained by alpha-compositing per-Gaussian normals defined as the thinnest axis of each Gaussian ellipsoid. The \textit{depth-derived normal} $\mathbf{n}_d$ is computed by back-projecting the rendered depth map $\mathbf{D}$ and taking finite-difference cross products. We also render a plane distance map $\mathbf{d}$, representing the projection of each point onto its local normal direction. These quantities are used by our DNGC regularization (Sec.~\ref{sec:dngc}).

\begin{figure*}[t]
\centering
\includegraphics[width=0.8\textwidth]{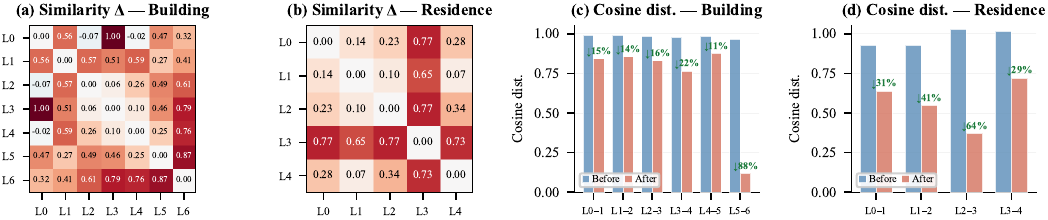}
\caption{\textbf{Quantitative analysis of CLCA on Building and Residence.}
\textbf{(a, b)} Change in inter-level cosine similarity after CLCA 
(warmer = larger increase). Both scenes show strong positive gains, 
particularly between fine levels (Building L5--L6: $\Delta{=}0.87$; 
Residence L2--L3: $\Delta{=}0.77$), confirming effective cross-level 
context flow.
\textbf{(c, d)} Adjacent-level cosine distance before and after CLCA. 
On Building, all pairs show 11--88\% reduction, with the finest pair 
(L5--L6) dropping from 0.97 to 0.12. On Residence, reductions range 
from 29\% to 64\% across all level pairs. More details are provided in the supplementary material.}
\label{fig:clca_analysis}
\end{figure*}

\subsection{Overview of HiCo-GS}
\label{sec:overview}

Given a set of posed images and a sparse SfM point cloud, our pipeline proceeds as follows (Fig.~\ref{fig:pipeline}). We first construct a multi-level octree from the point cloud, populating each voxel center with an anchor and its associated attributes (Sec.~\ref{sec:prelim}). For each training camera, we perform LOD-based visibility filtering via Eq.~\ref{eq:lod} followed by view-frustum culling to obtain the set of visible anchors.

The visible anchors' features $\{\mathbf{f}_i\}$ are then enhanced by our \textbf{Cross-Level Context Aggregation (CLCA)} module (Sec.~\ref{sec:clca}), which injects hierarchical context from parent and child octree levels into each anchor's feature to produce $\hat{\mathbf{f}}_i$. The enhanced features, concatenated with the viewing direction and optionally a per-camera appearance embedding, are fed into three shared MLPs to predict per-offset opacity, covariance (scale and rotation), and color. Offsets with positive opacity are retained and combined with anchor positions as $\mathbf{x}_i^k = \mathbf{p}_i + \boldsymbol{\delta}_i^k \odot \mathbf{s}_i^{[1:3]}$ to produce the final set of neural Gaussians.

Differentiable rasterization then produces the RGB image $\mathbf{I}$, rendered normal $\mathbf{n}_r$, depth map $\mathbf{D}$ (and its derived normal $\mathbf{n}_d$), plane distance $\mathbf{d}$, and opacity map $\boldsymbol{\alpha}$. The training loss combines photometric reconstruction with our \textbf{Depth-Normal Geometric Consistency (DNGC)} regularization (Sec.~\ref{sec:dngc}):
\begin{equation}
    \mathcal{L} = \mathcal{L}_{\text{rgb}} + w(t)\left(\lambda_{\text{nc}}\mathcal{L}_{\text{nc}} + \lambda_{\text{ns}}\mathcal{L}_{\text{ns}} + \lambda_{\text{ds}}\mathcal{L}_{\text{ds}}\right),
    \label{eq:total_loss}
\end{equation}
where $\mathcal{L}_{\text{rgb}}$ is the standard combination of $\ell_1$ and D-SSIM losses, and $w(t)$ is a linear warmup coefficient detailed in Sec.~\ref{sec:dngc}.

\subsection{Cross-Level Context Aggregation}
\label{sec:clca}

In existing octree-anchor methods, each level's anchor features are optimized independently through backpropagation. A fine-level anchor encoding a window frame has no knowledge of the coarse-level wall it belongs to, and a coarse-level anchor is unaware of the local texture complexity beneath it. We address this cross-level feature isolation through a module that enables bidirectional context flow across octree levels. Unlike multi-scale aggregation in point cloud networks~\cite{qi2017pointnet++} that extract hierarchical features from scratch in a single forward pass, CLCA injects cross-level priors into already-learned features during iterative rendering optimization. And unlike UNet-style skip connections across sequential encoder-decoder stages, our octree levels coexist simultaneously under LOD selection, requiring spatially-indexed lookup.

\noindent\textbf{Spatial-Aware Level Context.}
For each octree level $l$, we compute spatially-indexed context features. Given voxel size $v_l$ at level $l$, each visible anchor is mapped to its grid coordinate $\mathbf{g}_i^l = \lfloor (\mathbf{p}_i - \mathbf{o}) / v_l \rceil$, where $\mathbf{o}$ is the octree origin. Anchors at the same level sharing the same grid coordinate are aggregated via mean pooling:
\begin{equation}
    \mathbf{c}^l(\mathbf{g}) = \mathrm{MeanPool}\left(\left\{\mathbf{f}_i \;\middle|\; i \in \mathcal{V}_l,\; \mathbf{g}_i^l = \mathbf{g}\right\}\right),
    \label{eq:level_context}
\end{equation}
where $\mathcal{V}_l$ is the set of visible anchors at level $l$. 

\noindent\textbf{Hierarchical Context Lookup.}
The octree's spatial containment provides a natural parent-child relationship: an anchor at level $l$ resides within a specific voxel at level $l{-}1$. We compute each anchor's parent-level grid coordinate as $\mathbf{g}_i^{l-1} = \lfloor (\mathbf{p}_i - \mathbf{o}) / v_{l-1} \rceil$ and look up the corresponding context:
\begin{equation}
    \mathbf{c}_{\text{parent}}(\mathbf{a}_i) = \mathbf{c}^{l_i - 1}(\mathbf{g}_i^{l_i - 1}).
    \label{eq:parent_ctx}
\end{equation}
For child context, we aggregate all level-$(l{+}1)$ anchors mapping into the same level-$l$ grid cell as anchor $\mathbf{a}_i$:
\begin{equation}
    \mathbf{c}_{\text{child}}(\mathbf{a}_i) = \mathrm{MeanPool}\left(\left\{\mathbf{f}_j \;\middle|\; j \in \mathcal{V}_{l_i+1},\; \mathbf{g}_j^{l_i} = \mathbf{g}_i^{l_i}\right\}\right),
    \label{eq:child_ctx}
\end{equation}
\ie features of finer-level anchors whose parent-level grid coordinate matches that of $\mathbf{a}_i$. For anchors at the coarsest (or finest) active level, the self-level context is used as a fallback for the missing parent (or child). We limit scope to a one-hop triplet (parent-self-child): with $b{=}2$, adjacent levels span a $4\times$ resolution ratio, and more distant information propagates implicitly since each level's context is itself enriched by its neighbors.

The parent context $\mathbf{c}_{\text{parent}}$ carries \textit{structural priors}---coarse-level information about wall orientation, facade material, and building-scale color that informs fine-level anchors. The child context $\mathbf{c}_{\text{child}}$ provides \textit{detail statistics}---fine-level information about local texture density and micro-geometry that prevents coarse-level anchors from over-smoothing.

\noindent\textbf{Feature Enhancement.}
For each visible anchor $\mathbf{a}_i$ at level $l_i$ with self-context $\mathbf{c}_{\text{self}} = \mathbf{c}^{l_i}(\mathbf{g}_i^{l_i})$, we assemble the triplet and produce the enhanced feature via a lightweight MLP with residual connection:
\begin{equation}
    \hat{\mathbf{f}}_i = \mathbf{f}_i + \mathrm{MLP}_\theta\left([\mathbf{c}_{\text{parent}};\; \mathbf{c}_{\text{self}};\; \mathbf{c}_{\text{child}}]\right),
    \label{eq:clca}
\end{equation}
where $[\cdot;\cdot;\cdot]$ denotes concatenation and $\mathrm{MLP}_\theta$ consists of two linear layers with ReLU activation ($3d \to d \to d$, with $d{=}32$). The residual connection ensures that the module acts as a refinement: at initialization, the MLP outputs near-zero values and the representation reverts to the baseline behavior. The spatial lookup uses hash-based indexing with $O(N\log M)$ complexity, and the MLP adds only $3{,}168$ parameters---negligible relative to the main prediction MLPs. Empirical validation in Fig.~\ref{fig:clca_analysis}.

\subsection{Depth-Normal Geometric Consistency}
\label{sec:dngc}

The rasterization stage produces two independent normal estimates $\mathbf{n}_r$ and $\mathbf{n}_d$ (Sec.~\ref{sec:prelim}) that are derived from fundamentally different geometric information---Gaussian covariance and depth map, respectively. In existing methods, these two signals are not constrained to agree, allowing the Gaussian orientations to diverge from the actual surface geometry without penalty. This is particularly problematic in urban scenes where large planar regions (\eg building facades) should exhibit spatially coherent normals. We introduce three complementary regularization terms that leverage this geometric redundancy.

\noindent\textbf{Normal Consistency Loss.}
We enforce agreement between the two normal estimates, weighted by the accumulated opacity to focus on regions with sufficient Gaussian coverage:
\begin{equation}
    \mathcal{L}_{\text{nc}} = \frac{\sum_{\mathbf{p}} \boldsymbol{\alpha}(\mathbf{p}) \cdot \left\|\hat{\mathbf{n}}_r(\mathbf{p}) - \hat{\mathbf{n}}_d(\mathbf{p})\right\|_1}{\sum_{\mathbf{p}} \boldsymbol{\alpha}(\mathbf{p})},
    \label{eq:lnc}
\end{equation}
where $\hat{\mathbf{n}}_r$ and $\hat{\mathbf{n}}_d$ are $\ell_2$-normalized, and the summation is over all pixels $\mathbf{p}$.

\noindent\textbf{Edge-Aware Smoothness.}
Urban scenes exhibit a useful structural property: surface normals and depth should vary smoothly within planar regions (\eg a wall), while being allowed to change abruptly at object boundaries (\eg a window frame). We encode this prior through an edge-aware weighting derived from the RGB image gradient:
\begin{equation}
    w_x(\mathbf{p}) = \exp\left(-\beta \left|\nabla_x \mathbf{I}(\mathbf{p})\right|\right), \quad
    w_y(\mathbf{p}) = \exp\left(-\beta \left|\nabla_y \mathbf{I}(\mathbf{p})\right|\right),
    \label{eq:edge_weight}
\end{equation}
where $\nabla_x$ and $\nabla_y$ are horizontal and vertical finite differences of the rendered RGB image averaged across color channels, and $\beta=10$ controls edge sensitivity. This weight approaches 1 in flat regions (encouraging smoothness) and decays toward 0 at color edges (permitting discontinuities).

Using this weight, we define the normal smoothness and distance smoothness losses:
\begin{align}
    \mathcal{L}_{\text{ns}} &= \overline{w_x \cdot |\nabla_x \mathbf{n}_r|} + \overline{w_y \cdot |\nabla_y \mathbf{n}_r|}, \label{eq:lns} \\
    \mathcal{L}_{\text{ds}} &= \overline{w_x \cdot |\nabla_x \mathbf{d}|} + \overline{w_y \cdot |\nabla_y \mathbf{d}|}, \label{eq:lds}
\end{align}
where $\overline{(\cdot)}$ denotes the spatial mean. Both losses share the same edge weight $w$ from the RGB image: $\mathcal{L}_{\text{ns}}$ penalizes normal discontinuities on smooth surfaces (\eg a wall with an inconsistently oriented Gaussian), while $\mathcal{L}_{\text{ds}}$ penalizes distance jumps that indicate floating artifacts.

\noindent\textbf{Progressive Warmup.}
Geometric regularization requires reasonably converged depth and normal estimates; applying it too early when these signals are noisy can hinder RGB convergence. We therefore use a linear warmup:
\begin{equation}
    w(t) = \min\left(\frac{t - t_0}{T},\; 1\right),
    \label{eq:warmup}
\end{equation}
where $t_0$ is the iteration at which geometry rendering is activated and $T$ controls the ramp-up duration. This allows the photometric loss to dominate early training while geometric constraints are gradually introduced. We use $\lambda_{\text{nc}}=0.05$, $\lambda_{\text{ns}}=0.01$, $\lambda_{\text{ds}}=0.01$, and $T=10{,}000$ across all experiments.

\begin{figure}[t!]
    \centering
    \includegraphics[width=0.900\linewidth]{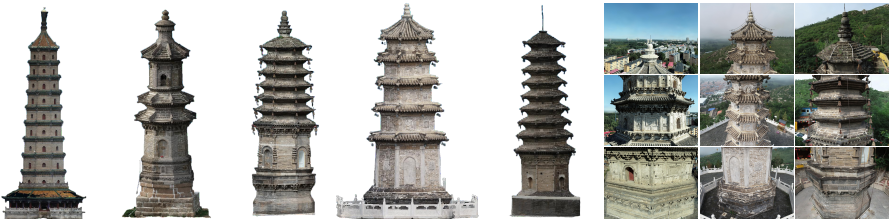}
    \caption{\textbf{Overview of the China-Pagoda dataset.} More details are provided in the supplementary material. }
    \label{fig:pagoda_gallery}
\end{figure}

\begin{table*}[t]
    \small
    \centering
    \caption{\textbf{Quantitative comparison on Mill19~\cite{turki2022mega} and UrbanScene3D~\cite{lin2022capturing}.} 
    $\uparrow$: higher is better, $\downarrow$: lower is better. 
    \first{Best} and \Second{second best} results are highlighted.
    $\dagger$: without decoupled appearance encoding.}
    \label{tab:compare}
    \resizebox{\linewidth}{!}{
        \large
        \begin{tabular}{l|ccc|ccc|ccc|ccc}
            \toprule
            \multirow{2}{*}{Method}
            & \multicolumn{3}{c|}{\emph{Building}}  
            & \multicolumn{3}{c|}{\emph{Rubble}} 
            & \multicolumn{3}{c|}{\emph{Residence}} 
            & \multicolumn{3}{c}{\emph{Sci-Art}} \\
            & SSIM$\uparrow$ & PSNR$\uparrow$ & LPIPS$\downarrow$   
            & SSIM$\uparrow$ & PSNR$\uparrow$ & LPIPS$\downarrow$ 
            & SSIM$\uparrow$ & PSNR$\uparrow$ & LPIPS$\downarrow$   
            & SSIM$\uparrow$ & PSNR$\uparrow$ & LPIPS$\downarrow$ \\
            \midrule
            \multicolumn{13}{l}{\textit{Without geometric optimization}} \\
            \midrule
            Mega-NeRF~\cite{turki2022mega}    
            & 0.547 & 20.92 & 0.454 
            & 0.553 & 24.06 & 0.508 
            & 0.628 & 22.08 & 0.401 
            & 0.770 & 25.60 & 0.312 \\
            Switch-NeRF~\cite{lin2022capturing}  
            & 0.579 & 21.54 & 0.397 
            & 0.562 & 24.31 & 0.478
            & 0.654 & \Second{22.57} & 0.352 
            & 0.795 & \Second{26.51} & 0.271 \\
            VastGaussian$\dagger$~\cite{lin2024vastgaussian}
            & 0.728 & 21.80 & 0.225 
            & 0.742 & 25.20 & 0.264 
            & 0.699 & 21.01 & 0.261 
            & 0.761 & 22.64 & 0.261 \\
            3DGS~\cite{kerbl20233d}
            & 0.738 & 22.53 & 0.214 
            & 0.725 & 25.51 & 0.316 
            & 0.745 & 22.36 & 0.247 
            & 0.791 & 24.13 & 0.262 \\ 
            DoGaussian~\cite{chen2024dogs}
            & 0.759 & 22.73 & 0.204
            & 0.765 & 25.78 & 0.257 
            & 0.740 & 21.94 & 0.244
            & 0.804 & 24.42 & 0.219 \\
            Momentum-GS~\cite{fan2025momentum}  
            & \Second{0.815} & \first{23.23} & \Second{0.194} 
            & \Second{0.827} & 25.93 & \Second{0.201} 
            & 0.818 & 22.21 & 0.197 
            & 0.856 & 23.02 & 0.205 \\
            CityGaussian~\cite{liu2024citygaussian}  
            & 0.778 & 21.55 & 0.246 
            & 0.813 & 25.77 & 0.228
            & 0.813 & 22.00 & 0.211
            & 0.837 & 21.39 & 0.230 \\
            \midrule
            \multicolumn{13}{l}{\textit{With geometric optimization}} \\
            \midrule
            SuGaR~\cite{guedon2024sugar}   
            & 0.507 & 17.76 & 0.455 
            & 0.577 & 20.69 & 0.453 
            & 0.603 & 18.74 & 0.406 
            & 0.698 & 18.60 & 0.349 \\
            NeuS~\cite{wang2021neus}     
            & 0.463 & 18.01 & 0.611 
            & 0.480 & 20.46 & 0.618 
            & 0.503 & 17.85 & 0.533 
            & 0.633 & 18.62 & 0.472 \\
            Neuralangelo~\cite{li2023neuralangelo}
            & 0.582 & 17.89 & 0.322
            & 0.625 & 20.18 & 0.314
            & 0.644 & 18.03 & 0.263
            & 0.769 & 19.10 & 0.231 \\
            PGSR~\cite{chen2024pgsr}
            & 0.480 & 16.12 & 0.573 
            & 0.728 & 23.09 & 0.334 
            & 0.746 & 20.57 & 0.289 
            & 0.799 & 19.72 & 0.275 \\
            PGSR+VastGS
            & 0.720 & 21.63 & 0.300
            & 0.768 & 25.32 & 0.274
            & -- & -- & --
            & -- & -- & -- \\
            CityGaussianV2~\cite{liu2024citygaussianv2}  
            & 0.650 & 19.07 & 0.397 
            & 0.720 & 23.75 & 0.322 
            & 0.769 & 21.15 & 0.234
            & 0.810 & 20.66 & 0.266 \\
            CityGS-$\mathcal{X}$~\cite{gao2025citygs}
            & \first{0.817} & \Second{22.76} & \first{0.191} 
            & 0.823 & \Second{26.15} & 0.210
            & \Second{0.819} & 22.44 & \Second{0.194} 
            & \Second{0.867} & 22.77 & \Second{0.179} \\
            \midrule
            \textbf{HiCo-GS (Ours)}  
            & {0.789} & {22.67} & {0.231} 
            & \first{0.863} & \first{28.09} & \first{0.192} 
            & \first{0.844} & \first{24.41} & \first{0.166} 
            & \first{0.894} & \first{26.64} & \first{0.163} \\
            \bottomrule
        \end{tabular}
    }
\end{table*}

\begin{table}[t]
    \centering
    \caption{\textbf{Comparison with existing 3D reconstruction benchmarks.} China-Pagoda provides dense per-scene coverage of geometrically extreme structures, complementing existing urban and cultural heritage datasets.}
    \label{tab:dataset_comparison}
    \resizebox{\linewidth}{!}{
    \begin{tabular}{lccccc}
        \toprule
        Dataset & Scenes & Total imgs & Imgs/scene & Type & Geometry \\
        \midrule
        Mill19~\cite{turki2022mega}           & 2  & $\sim$4K    & $\sim$2K   & Urban    & Planar facades \\
        UrbanScene3D~\cite{lin2022capturing} & 6  & $\sim$5K  & $\sim$0.8K & Urban    & Planar facades \\
        MatrixCity~\cite{li2023matrixcity}   & 2  & $\sim$10K   & $\sim$5K   & Synthetic & Mixed \\
        \midrule
        \textbf{China-Pagoda (Ours)}   & 8 & $\sim$10K+  & $\sim$1.2K+  & Heritage & Extreme \\
        \bottomrule
    \end{tabular}
    }
\end{table}

\subsection{China-Pagoda Benchmark}
\label{sec:dataset}
Existing urban reconstruction benchmarks such as Mill19~\cite{turki2022mega} and UrbanScene3D~\cite{lin2022capturing} are dominated by modern buildings with regular planar facades. Methods evaluated on these scenes can achieve high scores primarily by modeling flat surfaces well, while weaknesses in handling complex non-planar geometry remain unexposed. To provide a complementary evaluation axis, we introduce the \textbf{China-Pagoda} dataset, a collection of 8 ancient Chinese pagodas whose architectural characteristics differ fundamentally from modern urban structures.

\noindent\textbf{Data Collection.}
We capture 8 pagodas across northern China, spanning diverse styles from the Tang, Liao, and Qing dynasties (Fig.~\ref{fig:pagoda_gallery}). The structures range from 7-story brick pagodas with dense surface carvings to 13-story glazed-tile towers with elaborate bracket systems. For each pagoda, we collect over 1,200 raw images using drone-based aerial photography and ground-level captures, totaling over 10,000 images. Multi-altitude drone passes ensure complete coverage from base to spire, while ground-level captures provide close-range views of surface details.

\noindent\textbf{Geometric Challenges.}
The China-Pagoda scenes exhibit three characteristics that stress-test reconstruction methods: dense ornamental carvings with millimeter-level relief detail that push the resolution limits of anchor-based representations, curved multi-layer eaves with complex bracket systems (\textit{dougong}) whose non-planar, self-occluding geometry cannot be captured by simple planar assumptions, and repetitive fine-grained brick and tile textures that create strong visual ambiguity across viewpoints. The combination of intricate multi-scale geometry, non-planar self-occluding structures, and pervasive texture repetition exposes limitations that standard planar-dominated scenes fail to reveal.

Table~\ref{tab:dataset_comparison} compares China-Pagoda with existing benchmarks. While not the largest in total image count, it is uniquely positioned in per-scene density and geometric complexity. For each scene, we provide raw images, COLMAP-reconstructed camera poses and sparse point clouds, and a standard train/test split. The dataset will be publicly released.

\begin{figure*}[t]
    \centering
    \includegraphics[width=0.72\linewidth]{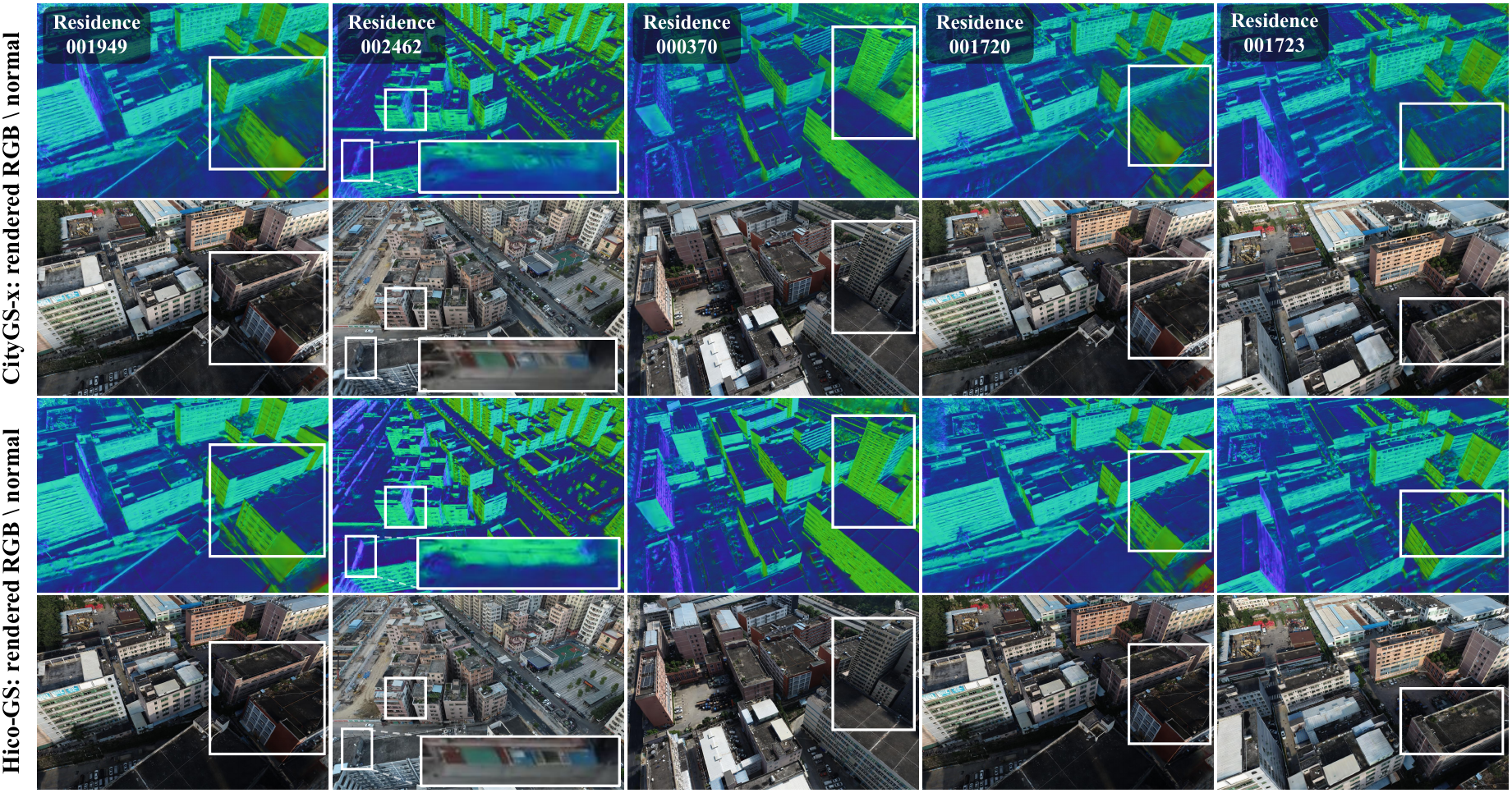}
    \caption{\textbf{Qualitative results of our method and other methods in large-scale reconstruction datasets UrbanScene.}}
    \label{fig:qual_render}
\end{figure*}

\begin{figure*}[t]
    \centering
    \includegraphics[width=0.72\linewidth]{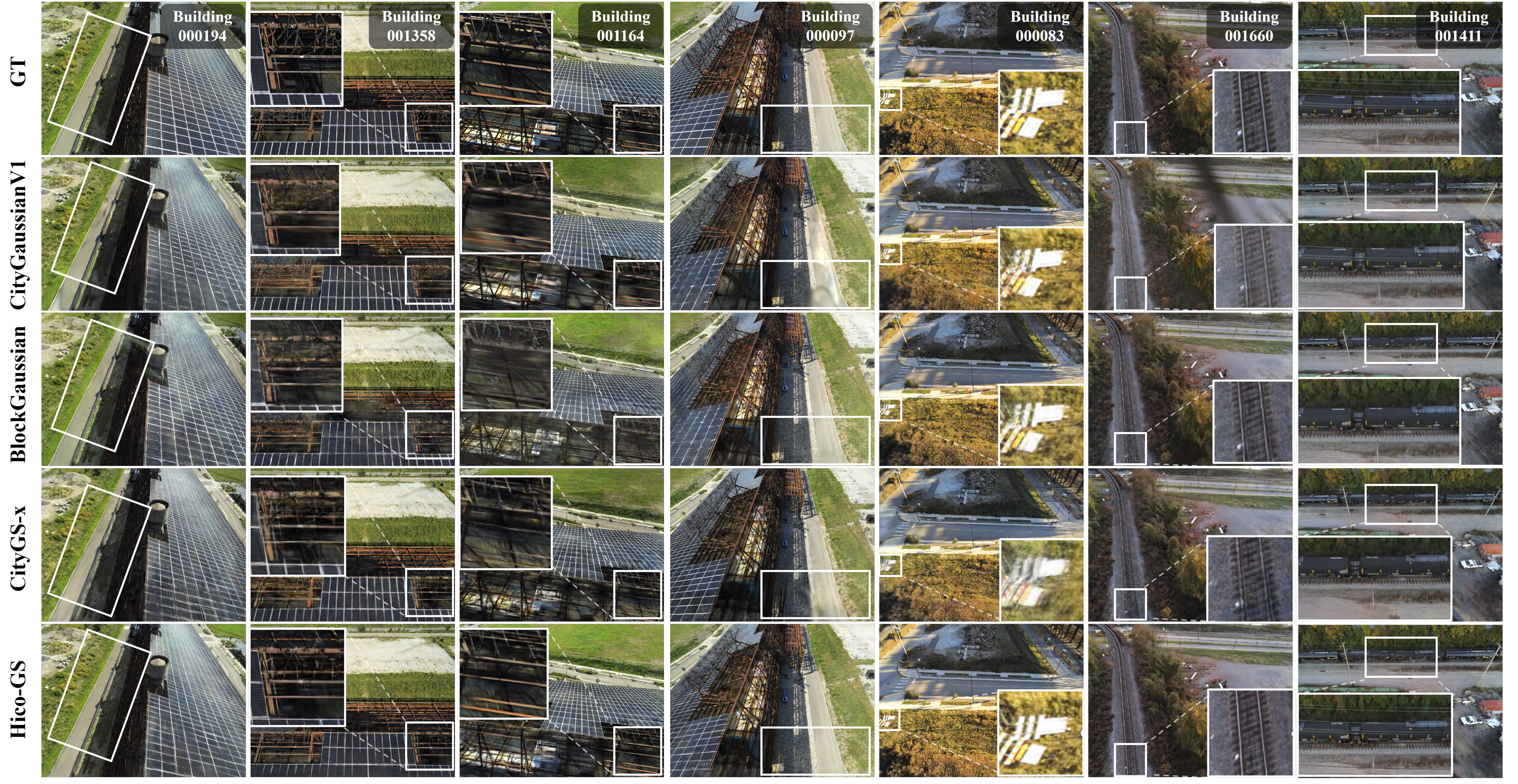}
    \caption{\textbf{Qualitative results of ours and other methods in image and depth rendering on Mill-19 and Urbanscene3D datasets.}}
    \label{fig:qual_normal}
\end{figure*}

\begin{figure}[t]
    \centering
    \includegraphics[width=0.7\linewidth]{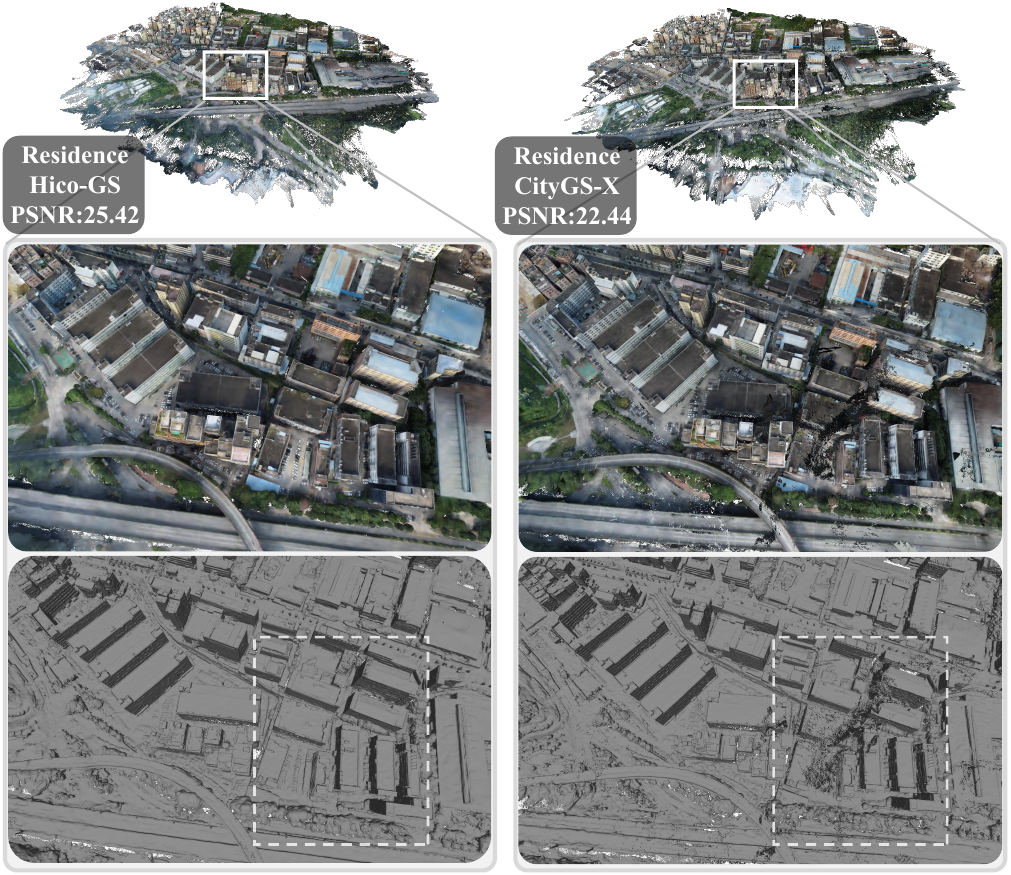}
    \caption{\textbf{Qualitative mesh and texture comparison between CityGS-$\mathcal{X}$ and our method on the Residence.}}
    \label{fig:qual_mesh}
\end{figure}

\begin{table*}[t]
    \centering
    \caption{\textbf{Results on China-Pagoda benchmark.} 
    $\uparrow$: higher is better, $\downarrow$: lower is better.
     }
    \label{tab:heritage}
    \resizebox{\linewidth}{!}{
        \large
        \begin{tabular}{l|ccc|ccc|ccc|ccc}
            \toprule
            \multirow{2}{*}{Method}
            & \multicolumn{3}{c|}{\emph{Baoanta}}  
            & \multicolumn{3}{c|}{\emph{Beita}} 
            & \multicolumn{3}{c|}{\emph{Duobaofota}} 
            & \multicolumn{3}{c}{\emph{Lingshan}} \\
            & SSIM$\uparrow$ & PSNR$\uparrow$ & LPIPS$\downarrow$   
            & SSIM$\uparrow$ & PSNR$\uparrow$ & LPIPS$\downarrow$ 
            & SSIM$\uparrow$ & PSNR$\uparrow$ & LPIPS$\downarrow$   
            & SSIM$\uparrow$ & PSNR$\uparrow$ & LPIPS$\downarrow$ \\
            \midrule
            CityGS-v1~\cite{liu2024citygaussian}    
            & 0.745 & 22.79 & 0.369 
            & 0.759 & \Second{23.92} & 0.427 
            & 0.806 & 23.52 & 0.308 
            & 0.879 & 25.31 & 0.211 \\
            CityGS-$\mathcal{X}$~\cite{gao2025citygs}  
            & \Second{0.891} & 26.33 & 0.168
            & \Second{0.770} & 23.17 & \first{0.350}
            & \Second{0.811} & 22.49 & \Second{0.217}
            & \Second{0.950} & \Second{28.77} & \Second{0.081} \\
            SplatCo~\cite{xiao2025splatco}
            & {0.827} & \Second{27.05} & \Second{0.273}
            & 0.643 & 21.29 & 0.578
            & 0.792 & \Second{25.88} & 0.304
            & 0.550 & 16.41 & 0.513 \\
            BlockGaussian~\cite{wu2025blockgaussian}
            & 0.760 & 23.48 & 0.349 
            & 0.686 & 21.74 & 0.459 
            & 0.700 & 20.29 & 0.415 
            & 0.816 & 25.06 & 0.269 \\
            Momentum-GS~\cite{fan2025momentum}  
            & 0.815 & 23.79 & 0.359 
            & 0.714 & 22.98 & 0.462 
            & 0.797 & 23.90 & 0.257
            & 0.921 & 27.24 & 0.160 \\
            \midrule
            \textbf{HiCo-GS (Ours)}  
            & \first{0.928} & \first{29.51} & \first{0.133} 
            & \first{0.770} & \first{24.01} & \Second{0.357}
            & \first{0.886} & \first{28.56} & \first{0.160} 
            & \first{0.965} & \first{32.95} & \first{0.066} \\
            \bottomrule
            \\[-0.6em]
            \toprule
            \multirow{2}{*}{Method}
            & \multicolumn{3}{c|}{\emph{Yingsi}}  
            & \multicolumn{3}{c|}{\emph{Yongwangta}} 
            & \multicolumn{3}{c|}{\emph{Yuhuangta}} 
            & \multicolumn{3}{c}{\emph{Yunjusi}} \\
            & SSIM$\uparrow$ & PSNR$\uparrow$ & LPIPS$\downarrow$   
            & SSIM$\uparrow$ & PSNR$\uparrow$ & LPIPS$\downarrow$ 
            & SSIM$\uparrow$ & PSNR$\uparrow$ & LPIPS$\downarrow$   
            & SSIM$\uparrow$ & PSNR$\uparrow$ & LPIPS$\downarrow$ \\
            \midrule
            CityGS-v1~\cite{liu2024citygaussian}    
            & \Second{0.674} & {21.54} & 0.448
            & 0.737 & {22.38} & 0.380
            & 0.783 & 22.91 & 0.324
            & 0.804 & 23.50 & 0.299 \\
            CityGS-$\mathcal{X}$~\cite{gao2025citygs}  
            & 0.612 & 17.22 & \Second{0.402}
            & \Second{0.824} & 21.97 & \Second{0.215}
            & \Second{0.865} & 24.03 & \Second{0.160}
            & \Second{0.892} & \Second{25.22} & \first{0.129} \\
            SplatCo~\cite{xiao2025splatco}
            & 0.640 & \Second{23.06} & 0.525
            & 0.771 & \Second{25.62} & 0.353
            & 0.855 & \Second{27.96} & 0.238
            & 0.801 & 23.88 & 0.273 \\
            BlockGaussian~\cite{wu2025blockgaussian}
            & 0.595 & 19.44 & 0.479 
            & 0.639 & 20.80 & 0.458 
            & 0.798 & {24.18} & 0.314
            & 0.704 & 20.12 & 0.382 \\
            Momentum-GS~\cite{fan2025momentum}  
            & 0.618 & 20.50 & 0.484 
            & 0.805 & 21.90 & 0.301
            & 0.743 & 22.03 & 0.352
            & 0.829 & 24.18 & 0.227 \\
            \midrule
            \textbf{HiCo-GS (Ours)}  
            & \first{0.813} & \first{24.82} & \first{0.264} 
            & \first{0.866} & \first{26.68} & \first{0.195}
            & \first{0.926} & \first{30.65} & \first{0.110} 
            & \first{0.903} & \first{26.82} & \Second{0.132} \\
            \bottomrule
        \end{tabular}
    }
\end{table*}

\section{Experiments}
\label{sec:experiments}

\subsection{Implementation Details}
\label{sec:impl}

We implement HiCo-GS on top of the CityGS-$\mathcal{X}$~\cite{gao2025citygs} codebase. All experiments are conducted on 4$\times$NVIDIA RTX 4090 GPUs with distributed training. We train for 30,000 iterations with a batch size equal to the number of GPUs. The CLCA module uses a two-layer MLP ($96 \to 32 \to 32$) with ReLU activation, sharing the same learning rate schedule as the color MLP. For DNGC, we set $\lambda_{\text{nc}}=0.05$, $\lambda_{\text{ns}}=0.01$, $\lambda_{\text{ds}}=0.01$, edge sensitivity $\beta=10$, and warmup duration $T=10{,}000$ iterations. The geometry rendering is activated at iteration $t_0$ following the same schedule as CityGS-$\mathcal{X}$. We use the default octree configuration with branching factor $b=2$, feature dimension $d=32$, and $K=5$ offsets per anchor. All other hyperparameters follow CityGS-$\mathcal{X}$.

\subsection{Main Results}
\label{sec:main_results}
 
\noindent\textbf{Novel View Synthesis.}
In Tab.~\ref{tab:compare} and Fig.~\ref{fig:qual_render}, we conduct both quantitative and qualitative comparisons on Mill19~\cite{turki2022mega} and UrbanScene3D~\cite{lin2022capturing} to evaluate the rendering quality of recent large-scale reconstruction methods \textit{w/} and \textit{w/o} geometric optimizations.
It is evident that HiCo-GS achieves state-of-the-art performance across all four scenes and all metrics, outperforming both categories of methods by a significant margin. Compared to our direct baseline CityGS-$\mathcal{X}$, we observe substantial improvements, \textit{e.g.}, a +1.94\,dB PSNR gain and a 0.018 LPIPS reduction on \textit{Rubble}, and a +3.87\,dB PSNR gain with a 0.016 LPIPS reduction on \textit{Sci-Art}. Notably, HiCo-GS also surpasses the best methods without geometric optimization: compared to Momentum-GS~\cite{fan2025momentum}, we achieve a +2.16\,dB PSNR improvement on \textit{Rubble} and a +2.20\,dB improvement on \textit{Residence}. These improvements are most pronounced on scenes with complex multi-scale structures, where cross-level context communication and geometric regularization are most beneficial.

 
\noindent\textbf{Surface Reconstruction.}
In Tab.~\ref{tab:matrixcity}, we compare our method with other surface reconstruction methods on MatrixCity~\cite{li2023matrixcity}. The experimental results demonstrate that HiCo-GS achieves state-of-the-art performance with a PSNR of 27.93\,dB and an F1 score of 0.599, surpassing CityGS-$\mathcal{X}$ by 0.35\,dB in PSNR and 0.018 in F1. The gains are more moderate than on real-world datasets, which is expected since MatrixCity's synthetic rendering produces perfectly consistent appearances across views, reducing the impact of cross-level context aggregation on appearance modeling.
 
Fig.~\ref{fig:qual_normal} presents a comparison between our method and other reconstruction approaches on UrbanScene3D. Our method yields notably smoother normals on planar wall surfaces while preserving sharp transitions at structural edges, validating the effectiveness of DNGC regularization. In contrast, baseline methods produce noisy and inconsistent normal maps, particularly in large planar regions such as building facades and rooftops.
Additionally, Fig.~\ref{fig:qual_mesh} presents qualitative mesh and texture comparisons between our method and CityGS-$\mathcal{X}$ on the \textit{Residence} scene. It can be observed that HiCo-GS produces more detailed and cleaner surface structures with fewer floating artifacts, closely resembling the actual geometry of the scene. In contrast, CityGS-$\mathcal{X}$ exhibits fragmented surfaces and geometric inaccuracies, particularly on building facades.

\begin{table}[t]
    \centering
    \caption{\textbf{Results on MatrixCity~\cite{li2023matrixcity}.} 
    \first{Best} and \Second{second best} results are highlighted.}
    \label{tab:matrixcity}
    \setlength{\tabcolsep}{12pt}
    \footnotesize
    \begin{tabular}{l|cccc}
        \toprule
        Method & PSNR$\uparrow$ & P$\uparrow$ & R$\uparrow$ & F1$\uparrow$ \\
        \midrule
        NeuS~\cite{wang2021neus}  
        & 16.76 & FAIL & FAIL & FAIL \\
        Neuralangelo~\cite{li2023neuralangelo}
        & 19.22 & 0.080 & 0.083 & 0.081 \\
        SuGaR~\cite{guedon2024sugar}
        & OOM & OOM & OOM & OOM \\
        GOF~\cite{yu2024gaussian}
        & 17.42 & FAIL & FAIL & FAIL \\ 
        2DGS~\cite{huang20242d}
        & 21.35 & 0.207 & 0.390 & 0.270 \\
        CityGS~\cite{liu2024citygaussian} 
        & 27.46 & 0.362 & 0.637 & 0.462 \\
        CityGS-V2~\cite{liu2024citygaussianv2}
        & 27.23 & 0.441 & 0.752 & 0.556 \\
        CityGS-$\mathcal{X}$~\cite{gao2025citygs}
        & \Second{27.58} & \Second{0.444} & \Second{0.840} & \Second{0.581} \\
        \midrule
        \textbf{HiCo-GS (Ours)}  
        & \first{27.93} & \first{0.461} & \first{0.856} & \first{0.599} \\
        \bottomrule
    \end{tabular}
\end{table}

\noindent\textbf{Evaluation on China-Pagoda.}
 In Tab.~\ref{tab:heritage}, we evaluate HiCo-GS on our proposed China-Pagoda benchmark to assess performance under extreme geometric complexity.
HiCo-GS outperforms all baselines across the majority of scenes and metrics, with improvements more substantial than on standard urban benchmarks. For example, on \textit{Yuhuangta} we achieve a +6.62\,dB PSNR gain and a 0.105 LPIPS reduction over CityGS-$\mathcal{X}$, and on \textit{Lingshan} we reach 32.95\,dB PSNR with an SSIM of 0.965. On \textit{Baoanta}, HiCo-GS improves upon the second-best method SplatCo~\cite{xiao2025splatco} by +2.19\,dB PSNR. Across all eight scenes, HiCo-GS achieves 22 out of 24 first-place rankings, with the remaining two on \textit{Beita} (SSIM and LPIPS) narrowly taken by CityGS-$\mathcal{X}$ (0.770 vs.\ 0.770 and 0.350 vs.\ 0.357).

\begin{table}[t]
    \centering
    \caption{\textbf{Ablation study.} PSNR (dB)$\uparrow$ on Mill19/UrbanScene3D.}
    \label{tab:ablation}
    \setlength{\tabcolsep}{4pt}
    \begin{tabular}{cc|cccc|c}
        \toprule
        CLCA & DNGC & Build. & Rub. & Res. & Sci. & Avg. Time \\
        \midrule
        -- & --
        & 22.40 & 20.57 & 21.81 & 22.04 & 2h44min \\
        Mean & --
        & {22.51} & \Second{26.29} & \Second{24.37} & \Second{26.43} & 2h51min \\
        -- & {\color{cmark}\checkmark}
        & \Second{22.66} & {25.75} & 23.24 & 24.39 & 2h44min \\
        Mean & {\color{cmark}\checkmark}
        & \first{22.67} & \first{28.09} & \first{24.41} & \first{26.64} & 2h51min \\
        Max/Attn & {\color{cmark}\checkmark}
        & 22.60 & 27.97 & 24.30 & 26.53 & 2h52min \\
        \bottomrule
    \end{tabular}
\end{table}

\noindent\textbf{Ablation Analysis.}
As shown in Table~\ref{tab:ablation}, both CLCA and DNGC independently yield substantial improvements over the baseline, with average PSNR gains of +3.20\,dB and +2.78\,dB respectively, confirming that the two modules address complementary failure modes. CLCA shows particularly strong gains on Rubble (+5.72\,dB) where multi-scale structural context is critical. DNGC delivers the largest single-scene improvement on Rubble (+5.18\,dB), reflecting its effectiveness in producing geometrically coherent surfaces. The full model achieves the best PSNR on all scenes, demonstrating complementary gains. Replacing mean pooling in CLCA with max or attention yields comparable or slightly lower PSNR, validating mean aggregation as a sufficient regional prior. Training overhead is negligible: the full model requires only 2 additional minutes over the baseline.

\section{Conclusion}
\label{sec:conclusion}
We presented HiCo-GS, a high-fidelity reconstruction framework that addresses two complementary limitations in octree-based urban scene reconstruction. CLCA resolves cross-level feature isolation by leveraging the octree's spatial containment structure for bidirectional hierarchical context flow, enriching fine-level features with structural priors and coarse-level features with detail statistics. DNGC enforces agreement between rendered and depth-derived normals through edge-aware smoothness losses, exploiting urban planar priors to suppress floating artifacts. We also introduced China-Pagoda, a benchmark of 8 ancient pagodas with over 30,000 images featuring dense ornamental details and repetitive textures. Experiments on Mill19, UrbanScene3D, MatrixCity, and China-Pagoda demonstrate state-of-the-art rendering quality and cleaner geometry across real-world and synthetic benchmarks.


\begin{acks}
This work was supported by the National Natural Science Foundation of China under Grants 62571437 and 62471394.
\end{acks}

\bibliographystyle{ACM-Reference-Format}
\bibliography{sample-base}










\clearpage
\setcounter{figure}{0}
\setcounter{table}{0}
\setcounter{equation}{0}
\renewcommand{\thefigure}{\arabic{figure}}
\renewcommand{\thetable}{\arabic{table}}
\renewcommand{\theequation}{\arabic{equation}}
\renewcommand{\thesubsection}{\Alph{subsection}}
\renewcommand{\thesection}{\Roman{section}}
\setcounter{section}{0}
\begin{center}
    {\LARGE\bfseries Supplementary File}
\end{center}
\vspace{0.5\baselineskip}
\section{Cross-Level Feature Incoherence Analysis}
\label{sec:appendix_feature_analysis}

We provide a comprehensive feature-space analysis to empirically validate the cross-level feature isolation problem identified in the main paper and to characterize how CLCA reshapes the learned representations. The analysis spans four urban scenes: Building (8 octree levels), Rubble (5 levels), Residence (6 levels), and SciArt (7 levels). All measurements are taken from fully converged models at 100K iterations.

\subsection{Cross-Level Cosine Distance Matrix}
\label{sec:appendix_cosine_distance}

For each model we extract all anchor features $\{\mathbf{f}_i\}$ and their octree levels $\{l_i\}$, then compute the mean pairwise cosine distance between every level pair $(l_a, l_b)$:
\begin{equation}
    d(l_a, l_b) = 1 - \frac{1}{|\mathcal{A}_a||\mathcal{A}_b|} \sum_{i \in \mathcal{A}_a} \sum_{j \in \mathcal{A}_b} \frac{\mathbf{f}_i^\top \mathbf{f}_j}{\|\mathbf{f}_i\| \|\mathbf{f}_j\|},
\end{equation}
where $\mathcal{A}_l = \{i : l_i = l\}$. A value of 1.0 indicates perfect orthogonality.

\noindent\textbf{Baseline.}
Across all four scenes (Figs.~\ref{fig:fa_building}--\ref{fig:fa_sciart}, panel~(a)), every off-diagonal entry falls within 0.94--1.00, confirming that baseline anchor features at different octree levels are nearly orthogonal regardless of their spatial relationship. The isolation worsens monotonically with level depth: in Building, adjacent-level distances increase from 0.949 (L0$\leftrightarrow$L1) to 0.999 (L6$\leftrightarrow$L7), indicating that the finest levels suffer the most severe isolation.

\noindent\textbf{After CLCA.}
Panels~(b) of the same images show that CLCA induces a block-diagonal structure. Fine-level features become highly coherent with one another (\eg L5$\leftrightarrow$L6 drops from 0.995 to 0.343 in Building, a 65.5\% reduction), while coarse-level features retain their original structure. This demonstrates a transition from unstructured orthogonality to hierarchically organized specialization.

\subsection{Adjacent-Level Distance Reduction}
\label{sec:appendix_adjacent_bar}

\begin{figure*}[t]
    \small
    \centering
    \subcaptionbox{Baseline\label{fig:fa_residence_base}}[0.27\textwidth]{%
        \includegraphics[width=\linewidth]{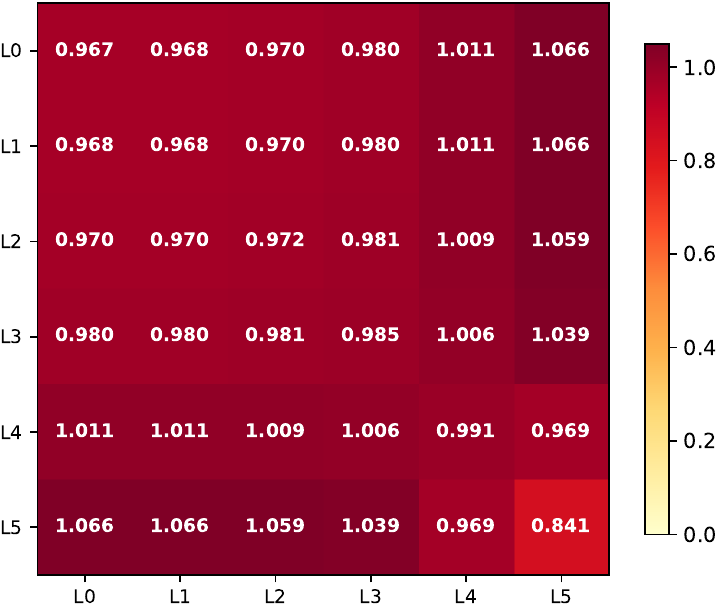}}\hfill
    \subcaptionbox{Ours (+CLCA)\label{fig:fa_residence_ours}}[0.27\textwidth]{%
        \includegraphics[width=\linewidth]{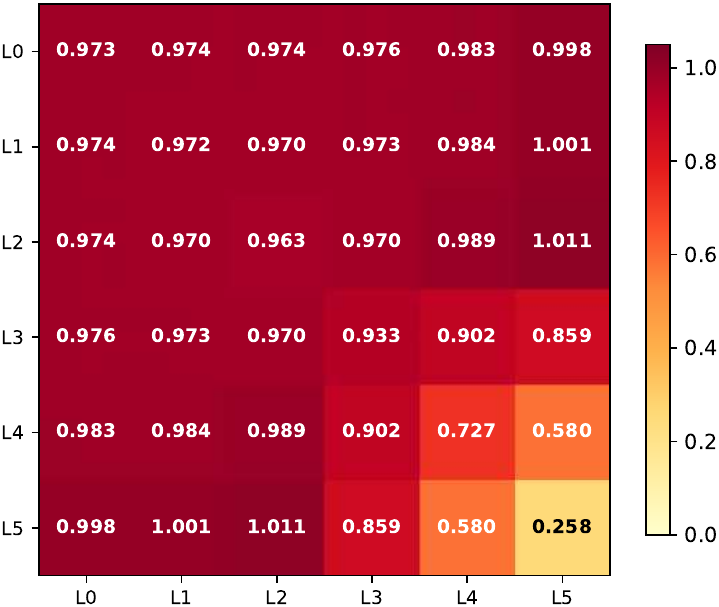}}\hfill
    \subcaptionbox{Adjacent-level distance comparison\label{fig:fa_residence_bar}}[0.40\textwidth]{%
        \includegraphics[width=\linewidth]{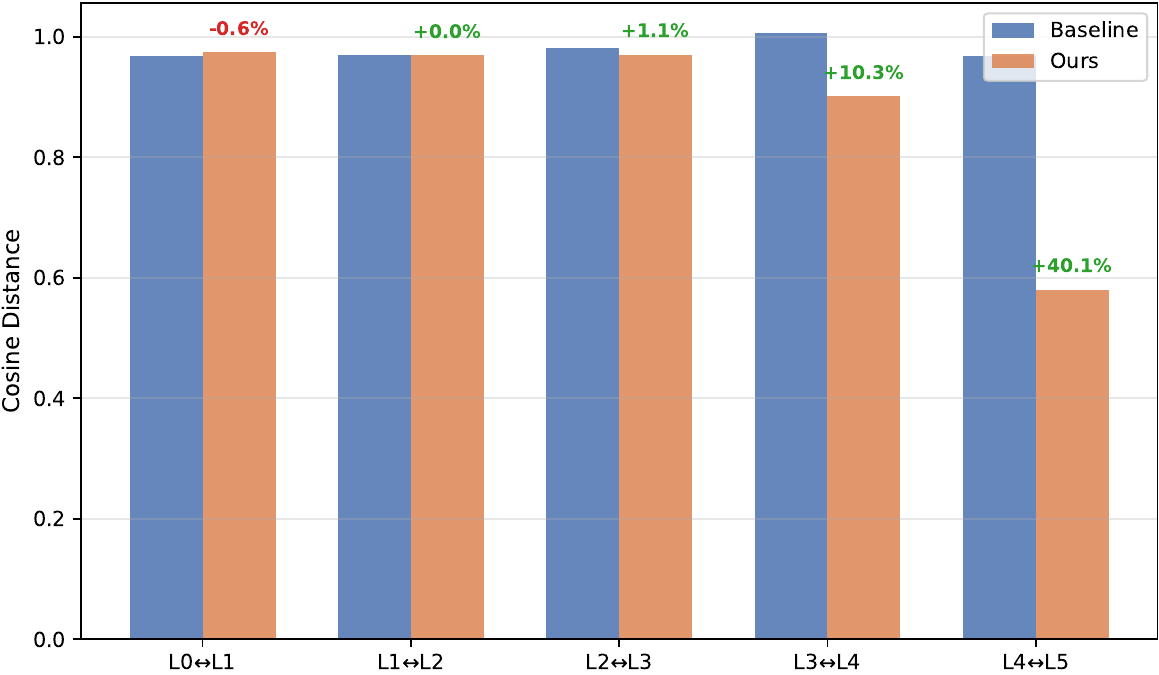}}
    \caption{\textbf{Cross-level feature analysis on the Residence scene (6 octree levels).}
    CLCA reduces adjacent-level distances by up to 40.1\% at the finest pair (L4$\leftrightarrow$L5), with coarse pairs showing minimal change. The adaptive coupling pattern is consistent with the other scenes.}
    \label{fig:fa_residence}
\end{figure*}

\begin{figure*}[t]
    \small
    \centering
    \subcaptionbox{Baseline\label{fig:fa_sciart_base}}[0.27\textwidth]{%
        \includegraphics[width=\linewidth]{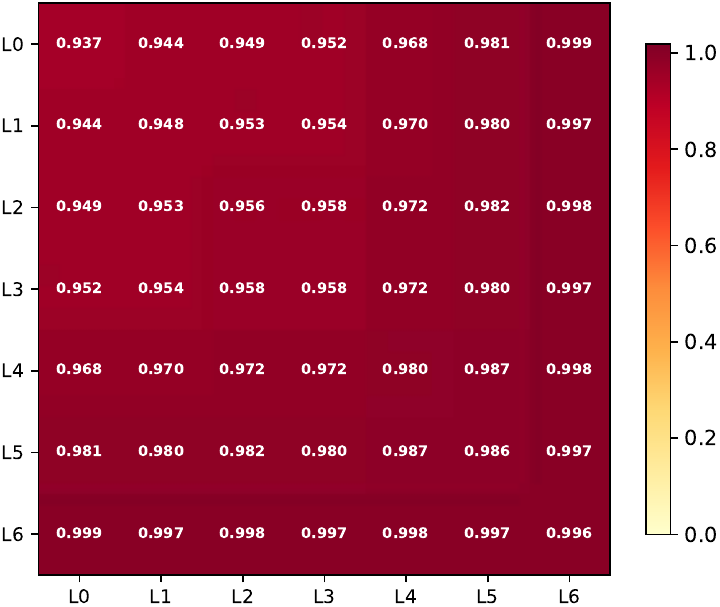}}\hfill
    \subcaptionbox{Ours (+CLCA)\label{fig:fa_sciart_ours}}[0.27\textwidth]{%
        \includegraphics[width=\linewidth]{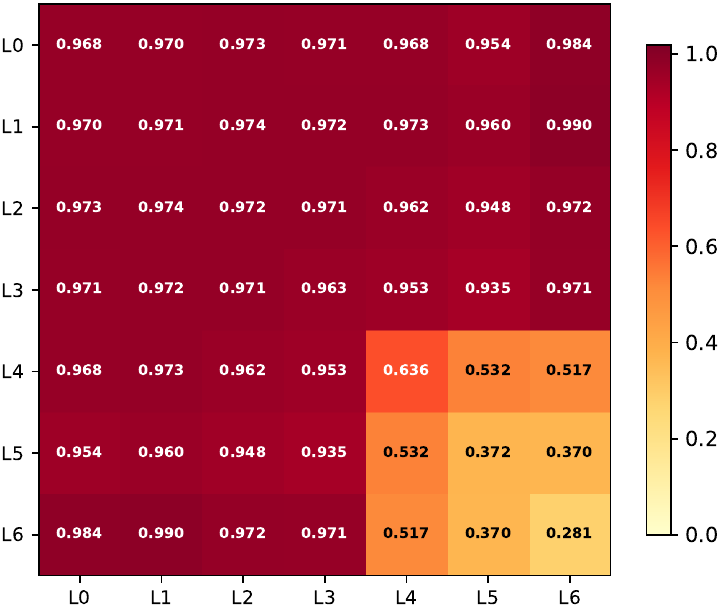}}\hfill
    \subcaptionbox{Adjacent-level distance comparison\label{fig:fa_sciart_bar}}[0.40\textwidth]{%
        \includegraphics[width=\linewidth]{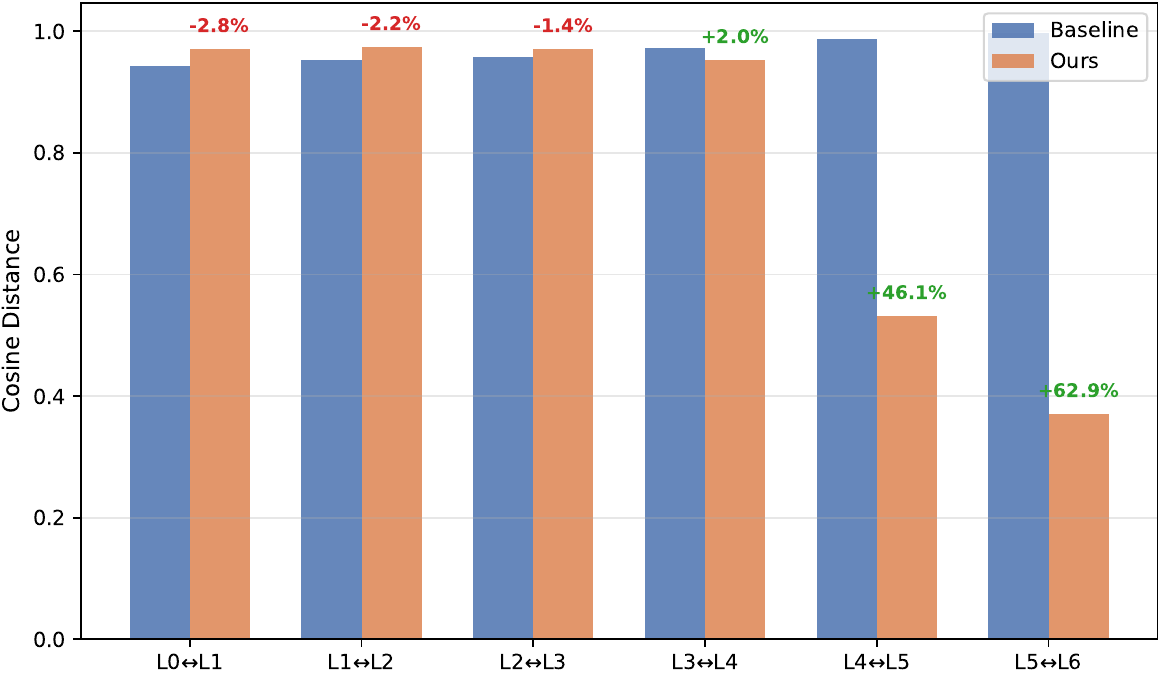}}
    \caption{\textbf{Cross-level feature analysis on the SciArt scene (7 octree levels).}
    SciArt exhibits the largest fine-level reductions among all scenes (L5$\leftrightarrow$L6: $-$62.9\%), consistent with its higher geometric complexity requiring stronger cross-level communication.}
    \label{fig:fa_sciart}
\end{figure*}

Panels~(c) of Figs.~\ref{fig:fa_building}--\ref{fig:fa_sciart} compare adjacent-level cosine distances before and after CLCA. A consistent pattern emerges across all scenes: the reduction concentrates overwhelmingly on the finest level pairs.

\begin{itemize}[leftmargin=*,nosep]
    \item \textbf{Building} (8 levels): L0$\leftrightarrow$L1 through L4$\leftrightarrow$L5 change by $\leq$3.9\%, while L5$\leftrightarrow$L6 and L6$\leftrightarrow$L7 are reduced by 65.5\% and 25.0\%.
    \item \textbf{Rubble} (5 levels): L3$\leftrightarrow$L4 is reduced by 28.8\%.
    \item \textbf{Residence} (6 levels): L3$\leftrightarrow$L4 and L4$\leftrightarrow$L5 are reduced by 10.3\% and 40.1\%.
    \item \textbf{SciArt} (7 levels): L4$\leftrightarrow$L5 and L5$\leftrightarrow$L6 are reduced by 46.1\% and 62.9\%.
\end{itemize}

This adaptive behavior is not manually designed but emerges from training dynamics. Coarse levels contain many anchors whose mean-pooled context approximates the global average and carries little discriminative information. Fine levels contain few anchors with distinctive statistics; the parent-level context provides substantial structural information that independent optimization cannot access, producing large feature adjustments. The network thus automatically concentrates coupling effort where incoherence is most severe.

\subsection{Per-Pair PCA Analysis}
\label{sec:appendix_perpair_pca}

Global dimensionality reduction is dominated by coarse-level anchors, which vastly outnumber fine-level ones. To reveal fine-grained cross-level structure, we perform per-pair PCA: for each adjacent pair $(l, l{+}1)$, we extract only anchors at these two levels (subsampled to 3{,}000 per level), fit PCA on their combined features, and project to 2D. The cosine similarity between the two levels' mean feature vectors is annotated.

\begin{figure*}[t]
    \small
    \centering
    \subcaptionbox{Baseline\label{fig:fa_building_base}}[0.27\textwidth]{%
        \includegraphics[width=\linewidth]{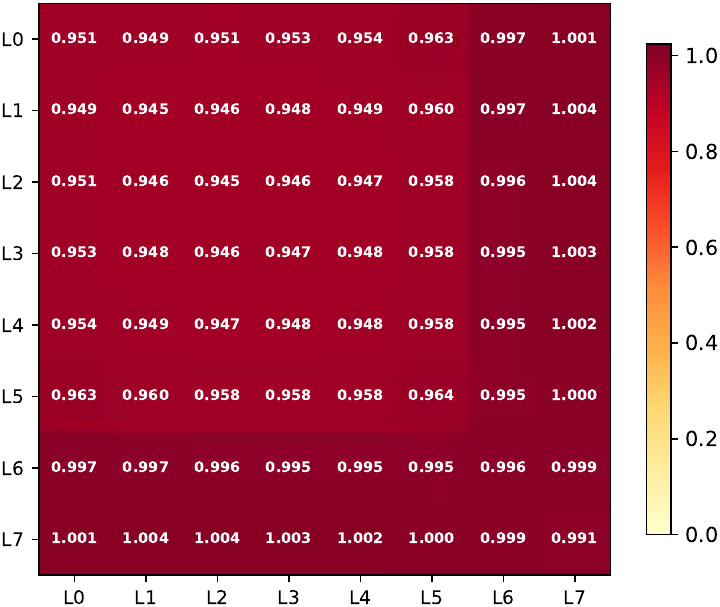}}\hfill
    \subcaptionbox{Ours (+CLCA)\label{fig:fa_building_ours}}[0.27\textwidth]{%
        \includegraphics[width=\linewidth]{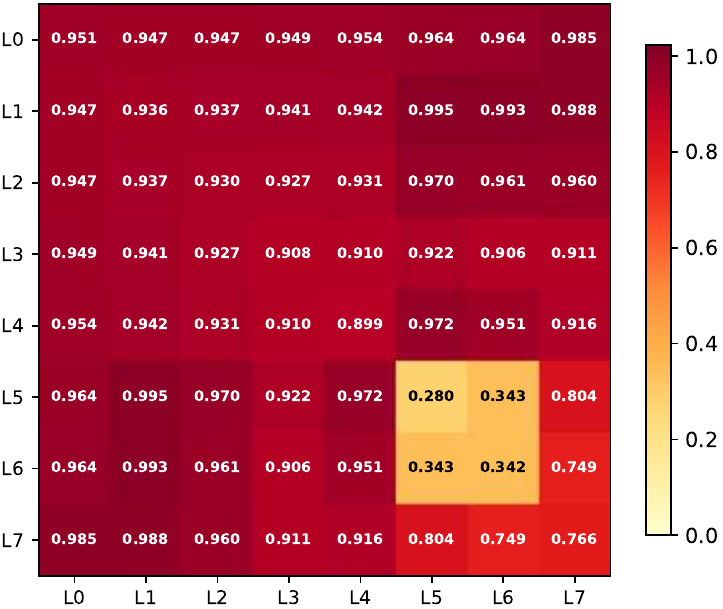}}\hfill
    \subcaptionbox{Adjacent-level distance comparison\label{fig:fa_building_bar}}[0.40\textwidth]{%
        \includegraphics[width=\linewidth]{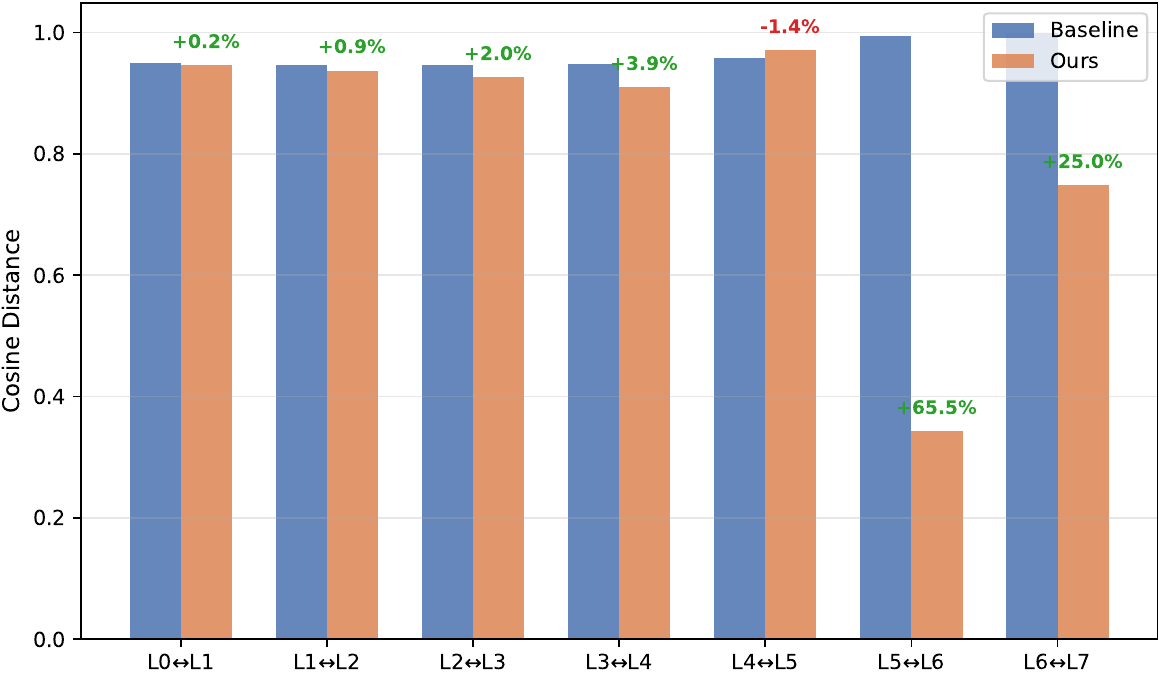}}

    \vspace{10pt}
    \subcaptionbox{Per-pair PCA projections of adjacent-level features\label{fig:fa_building_pca}}[0.98\textwidth]{%
        \includegraphics[width=\linewidth]{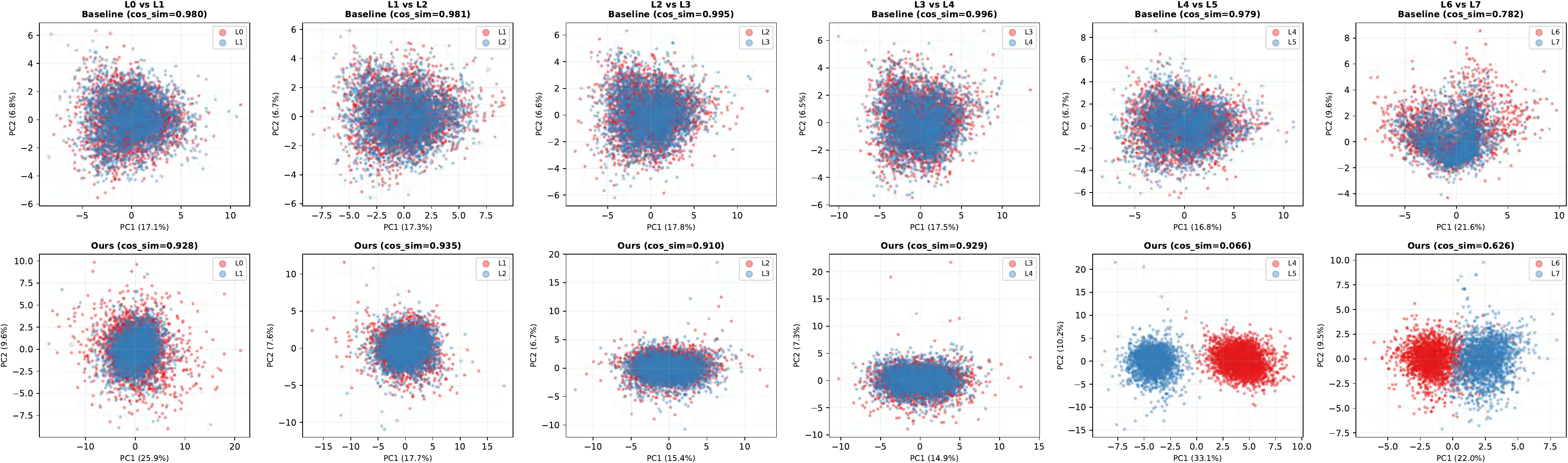}}
    \caption{\textbf{Cross-level feature analysis on the Building scene (8 octree levels, 5.98M anchors).}
    (a)~Baseline cosine distance matrix: all off-diagonal entries fall within 0.94--1.00, with incoherence worsening at finer levels (L6$\leftrightarrow$L7: 0.999).
    (b)~After CLCA: a block-diagonal structure emerges, with fine-level coherence dramatically improved (L5$\leftrightarrow$L6: 0.995$\to$0.343).
    (c)~Adjacent-level distance reductions concentrate on the finest pairs (L5$\leftrightarrow$L6: $-$65.5\%, L6$\leftrightarrow$L7: $-$25.0\%), while coarse pairs change by $\leq$4\%.
    (d)~Per-pair PCA: baseline features (top) are diffusely mixed across levels; after CLCA (bottom), fine-level pairs form clearly separated clusters with increased PC1 explained variance, indicating emergent scale-discriminative structure.}
    \label{fig:fa_building}
\end{figure*}

\begin{figure*}[t]
    \small
    \centering
    \subcaptionbox{Baseline\label{fig:fa_rubble_base}}[0.27\textwidth]{%
        \includegraphics[width=\linewidth]{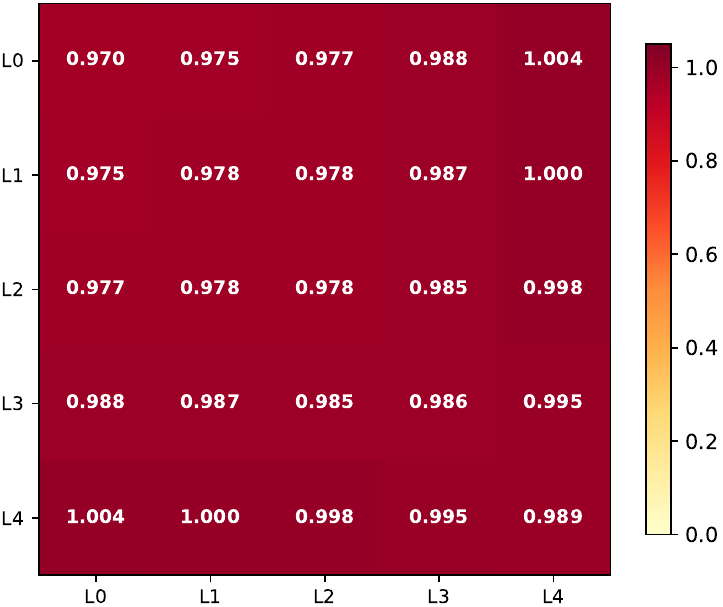}}\hfill
    \subcaptionbox{Ours (+CLCA)\label{fig:fa_rubble_ours}}[0.27\textwidth]{%
        \includegraphics[width=\linewidth]{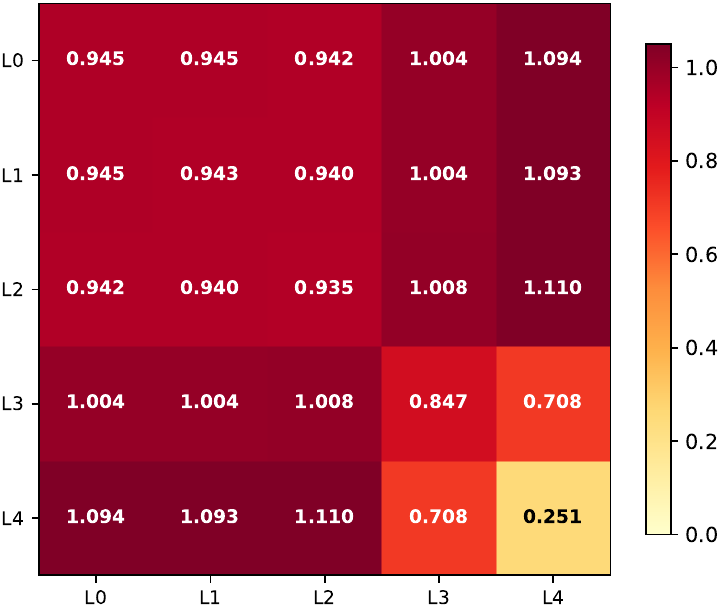}}\hfill
    \subcaptionbox{Adjacent-level distance comparison\label{fig:fa_rubble_bar}}[0.40\textwidth]{%
        \includegraphics[width=\linewidth]{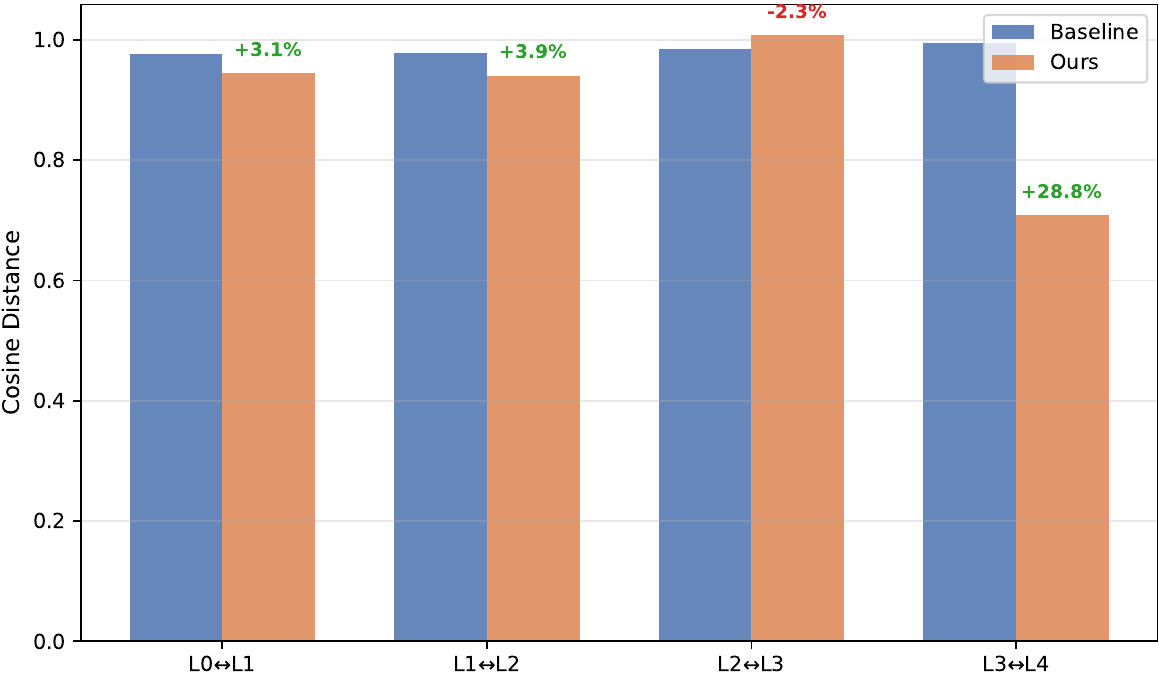}}

    \vspace{4pt}
    \subcaptionbox{Per-pair PCA projections of adjacent-level features\label{fig:fa_rubble_pca}}[0.98\textwidth]{%
        \includegraphics[width=\linewidth]{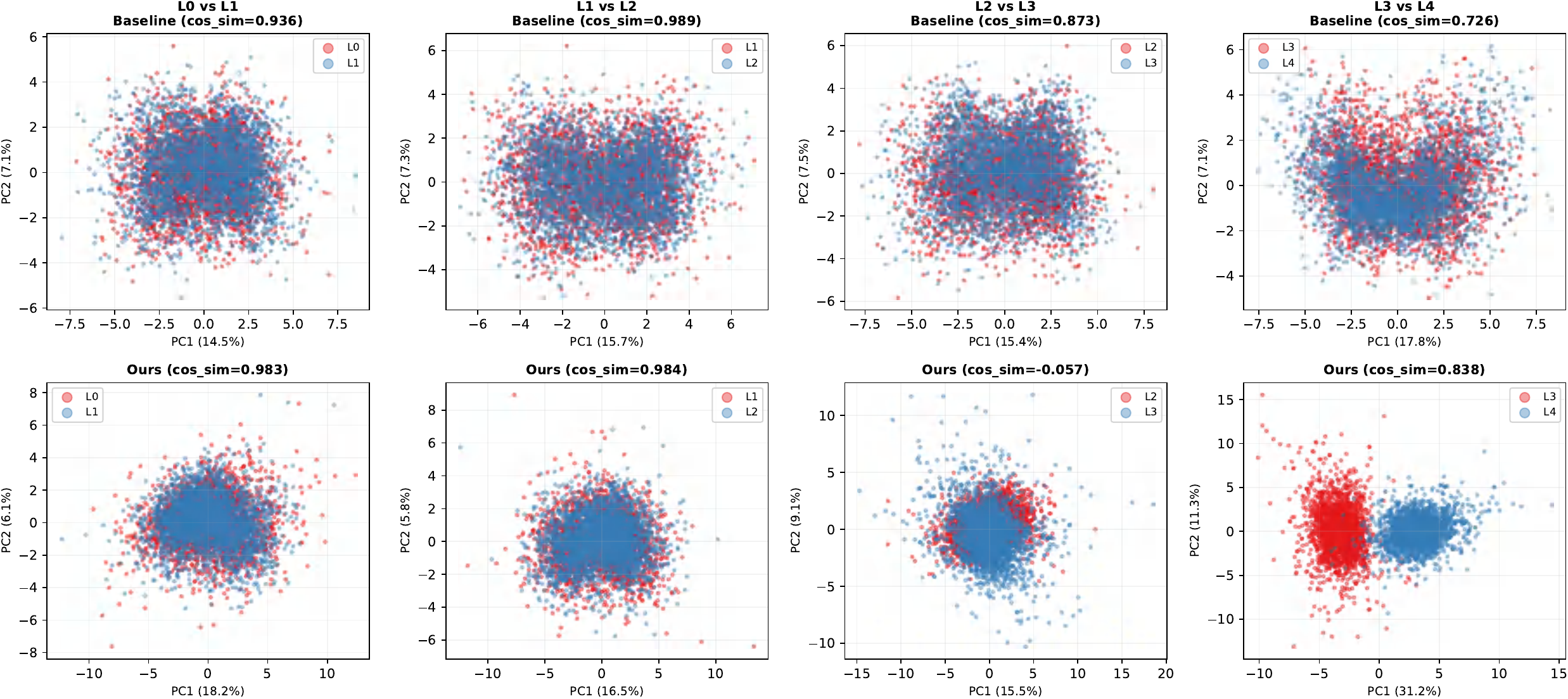}}
    \caption{\textbf{Cross-level feature analysis on the Rubble scene (5 octree levels).}
    The same analysis protocol as Fig.~\ref{fig:fa_building}. The baseline exhibits uniform near-orthogonality across all level pairs. After CLCA, the finest pair (L3$\leftrightarrow$L4) shows a 28.8\% distance reduction, while coarse pairs remain stable. Per-pair PCA confirms that CLCA selectively organizes fine-level features into separable clusters without disrupting coarse-level representations.}
    \label{fig:fa_rubble}
\end{figure*}

Fig.~\ref{fig:fa_building}(d) shows the Building scene results. In the baseline (top row), most pairs exhibit  overlapping point clouds (cos\_sim $\geq$ 0.78), confirming diffuse distributions without level-specific organization. After CLCA (bottom row), three distinct regimes appear: (i)~coarse pairs remain unchanged; (ii)~at the coarse-to-fine transition (\eg L4 vs L5), cos\_sim drops sharply from 0.979 to 0.066 and the two levels form clearly separated clusters, with PC1 explained variance rising from 16.8\% to 33.1\%, indicating that the principal axis of variation has become a scale-discriminative direction; (iii)~at the finest pairs (\eg L6 vs L7), both levels form compact, well-separated groups. This progression directly visualizes how CLCA induces functional specialization at fine octree levels.


\subsection{t-SNE Feature Visualization}
\label{sec:appendix_tsne}

\begin{figure*}[t]
    \centering
    \subcaptionbox{Building (8 levels, 1{,}500 per level)\label{fig:tsne_building}}[0.48\textwidth]{%
        \includegraphics[width=\linewidth]{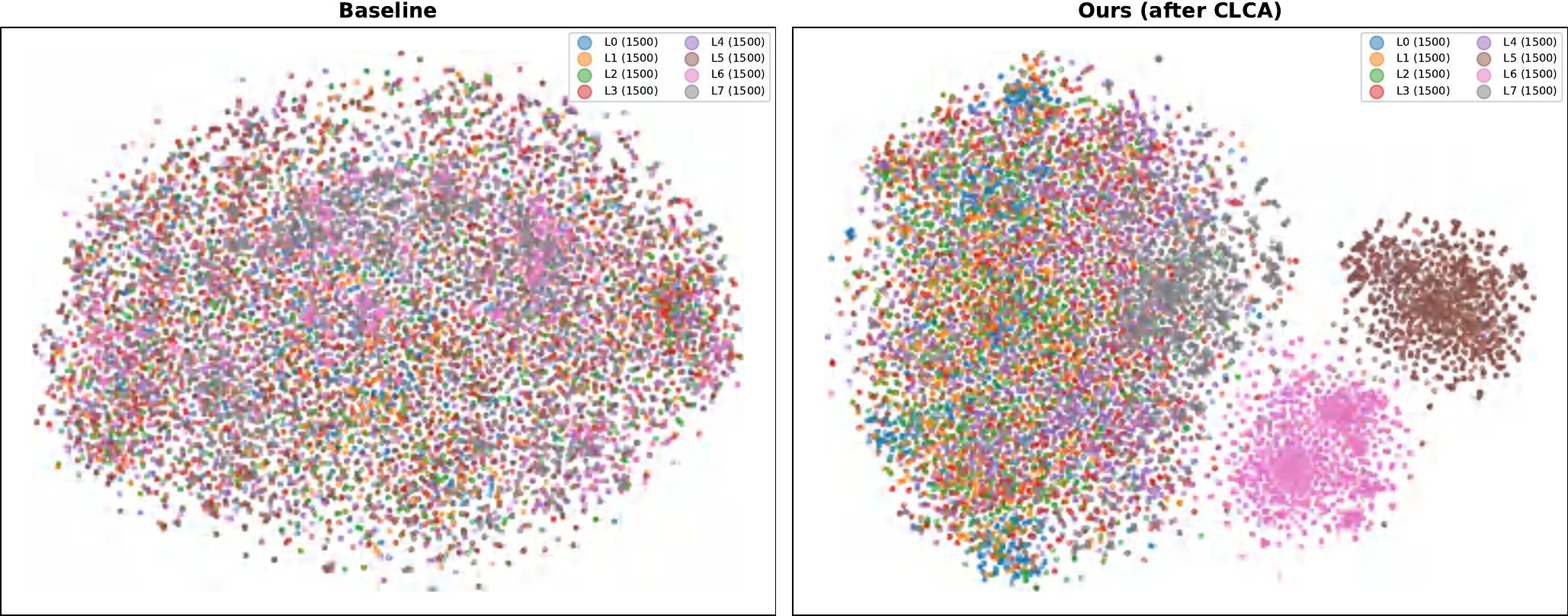}}\hfill
    \subcaptionbox{Rubble (5 levels, 2{,}400 per level)\label{fig:tsne_rubble}}[0.48\textwidth]{%
        \includegraphics[width=\linewidth]{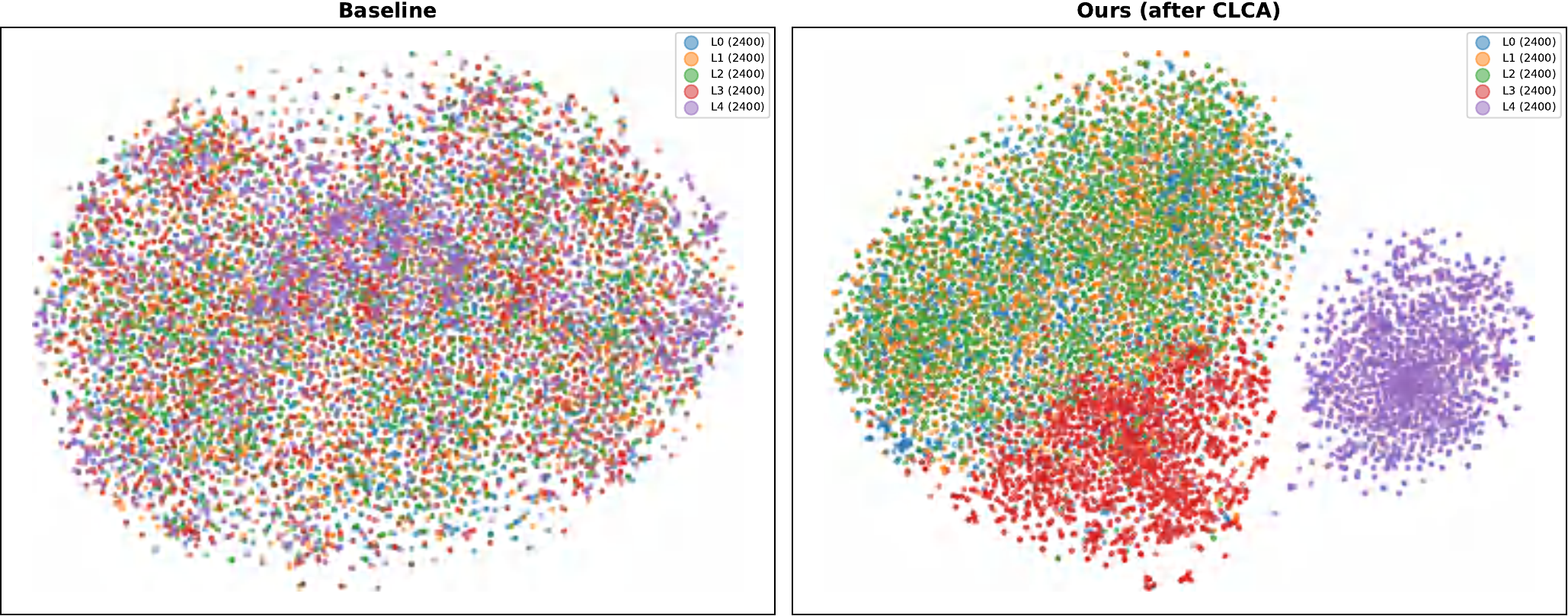}}

    \vspace{4pt}

    \subcaptionbox{Residence (6 levels, 2{,}000 per level)\label{fig:tsne_residence}}[0.48\textwidth]{%
        \includegraphics[width=\linewidth]{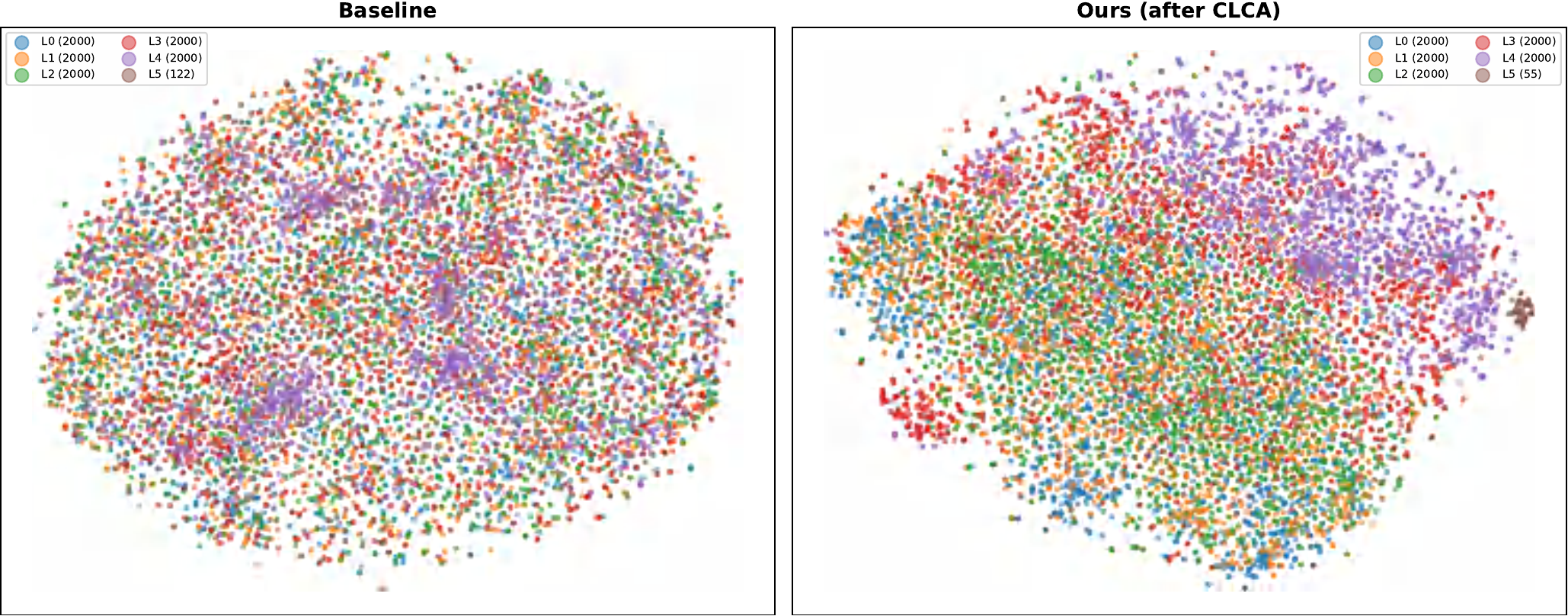}}\hfill
    \subcaptionbox{SciArt (7 levels, 1{,}714 per level)\label{fig:tsne_sciart}}[0.48\textwidth]{%
        \includegraphics[width=\linewidth]{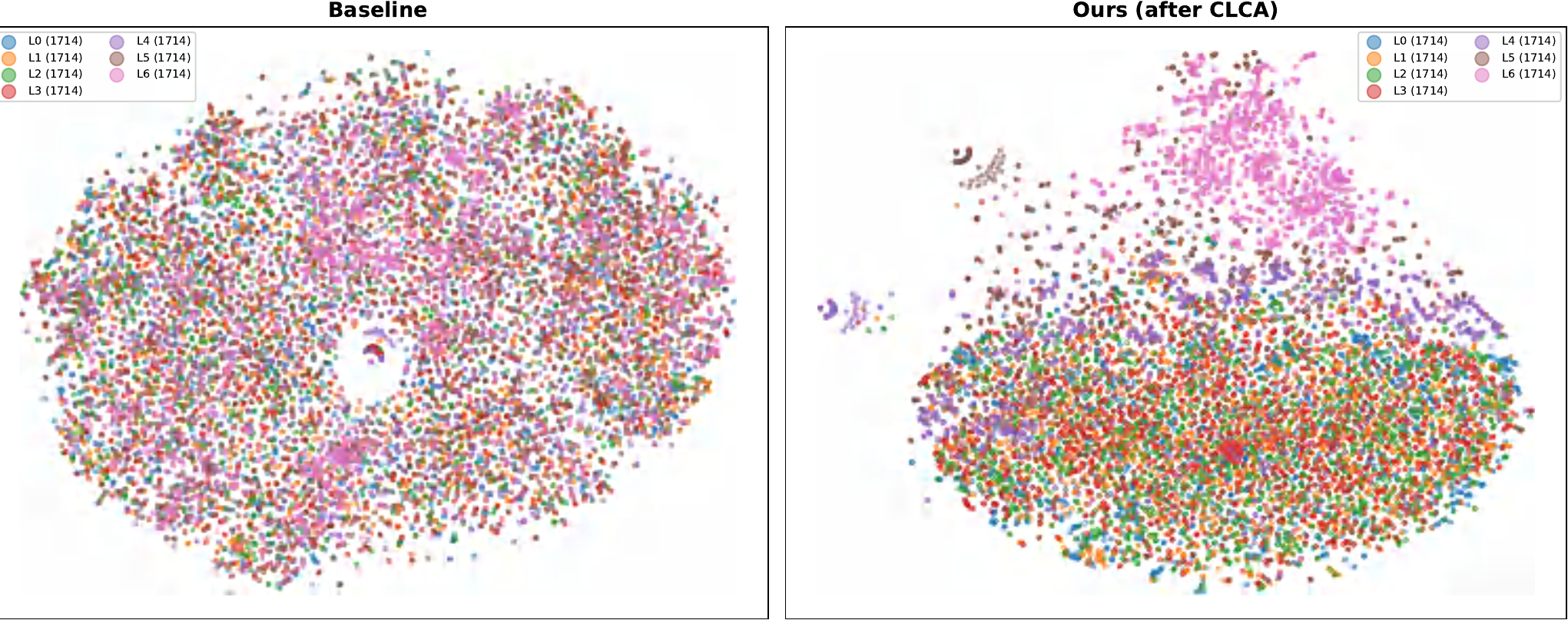}}
    \caption{\textbf{t-SNE visualization of anchor features across all octree levels.} Each panel shows the baseline (\textit{left}) and after CLCA (\textit{right}) for one scene, with stratified sampling ensuring equal representation per level. In the baseline, all levels form a single unstructured mixture. After CLCA, fine-level features consistently separate into distinct peripheral clusters while coarse levels share a common core, demonstrating the emergent hierarchical specialization induced by cross-level context aggregation.}
    \label{fig:tsne_all}
\end{figure*}

To provide a holistic view of the entire feature space, we perform t-SNE on all levels simultaneously for each scene. Since fine levels contain far fewer anchors than coarse levels, we adopt stratified sampling to ensure adequate representation of every level. We use a perplexity of 30 and run for 1{,}000 iterations. Fig.~\ref{fig:tsne_all} presents the results across all four scenes.

In the baseline (left panels), every scene exhibits a single unstructured blob in which all levels are uniformly intermixed. No level-specific clustering is visible, confirming the absence of organized inter-level structure: the near-orthogonality revealed by the cosine distance matrices manifests as a diffuse, undifferentiated embedding.

After CLCA (right panels), the feature space self-organizes into a hierarchically structured layout. Coarse-level anchors (shown in cooler tones) remain intermixed with each other, forming a shared core that encodes global scene structure. Fine-level anchors, in contrast, separate into distinct, tight clusters at the periphery of the embedding. This pattern is strikingly consistent across scenes of varying depth and complexity:

\begin{itemize}[leftmargin=*,nosep]
    \item \textbf{Building} (8 levels): L5, L6, and L7 each form clearly isolated satellite clusters around the coarse-level core (L0--L4).
    \item \textbf{Rubble} (5 levels): L3 and L4 pull away from the main body, with L4 forming the most compact cluster.
    \item \textbf{Residence} (6 levels): L5 separates as a small but distinct group, while L4 begins to detach from the coarse-level mass.
    \item \textbf{SciArt} (7 levels): L5 and L6 form peripheral clusters, with the remaining levels sharing a common central region.
\end{itemize}

This hierarchical organization emerges purely from the rendering loss. CLCA provides fine-level anchors with sufficient shared context to form coherent groups, while their separation from coarse levels gives the shared MLP decoder a clear signal for scale-dependent attribute prediction. The transition from an unstructured blob to a hierarchically organized embedding is the most direct visual evidence that CLCA transforms the feature space from uninformative orthogonality into functional specialization.

\renewcommand{\thesection}{\Roman{section}}
\setcounter{section}{1}
\section{Additional DNGC Ablation Results}
\label{sec:app_dngc}

\begin{figure*}[p]
    \centering
    \includegraphics[width=0.70\linewidth]{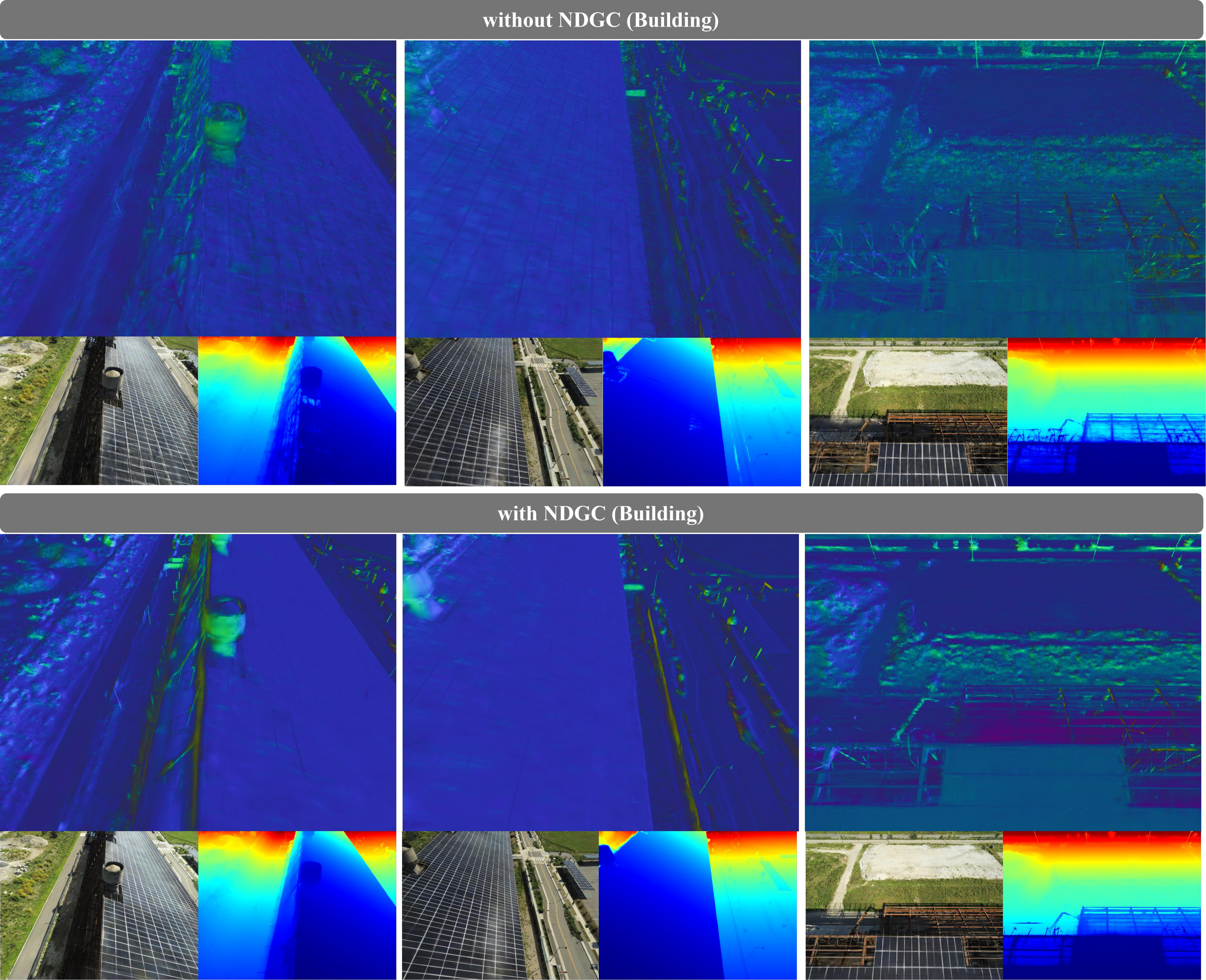}
    \caption{\textbf{DNGC ablation on the Building scene.}
    \textit{Top:} With DNGC, rendered normals exhibit clean, consistent orientations on planar surfaces such as rooftops and walls, and depth maps show smooth gradients with sharp object boundaries.
    \textit{Bottom:} Without DNGC, normals are noticeably noisier with inconsistent directions on the same regions, and depth maps contain discontinuities and floating artifacts around building facades and ground planes.}
    \label{fig:app_dngc_building}

    \vspace{4pt}

    \includegraphics[width=0.70\linewidth]{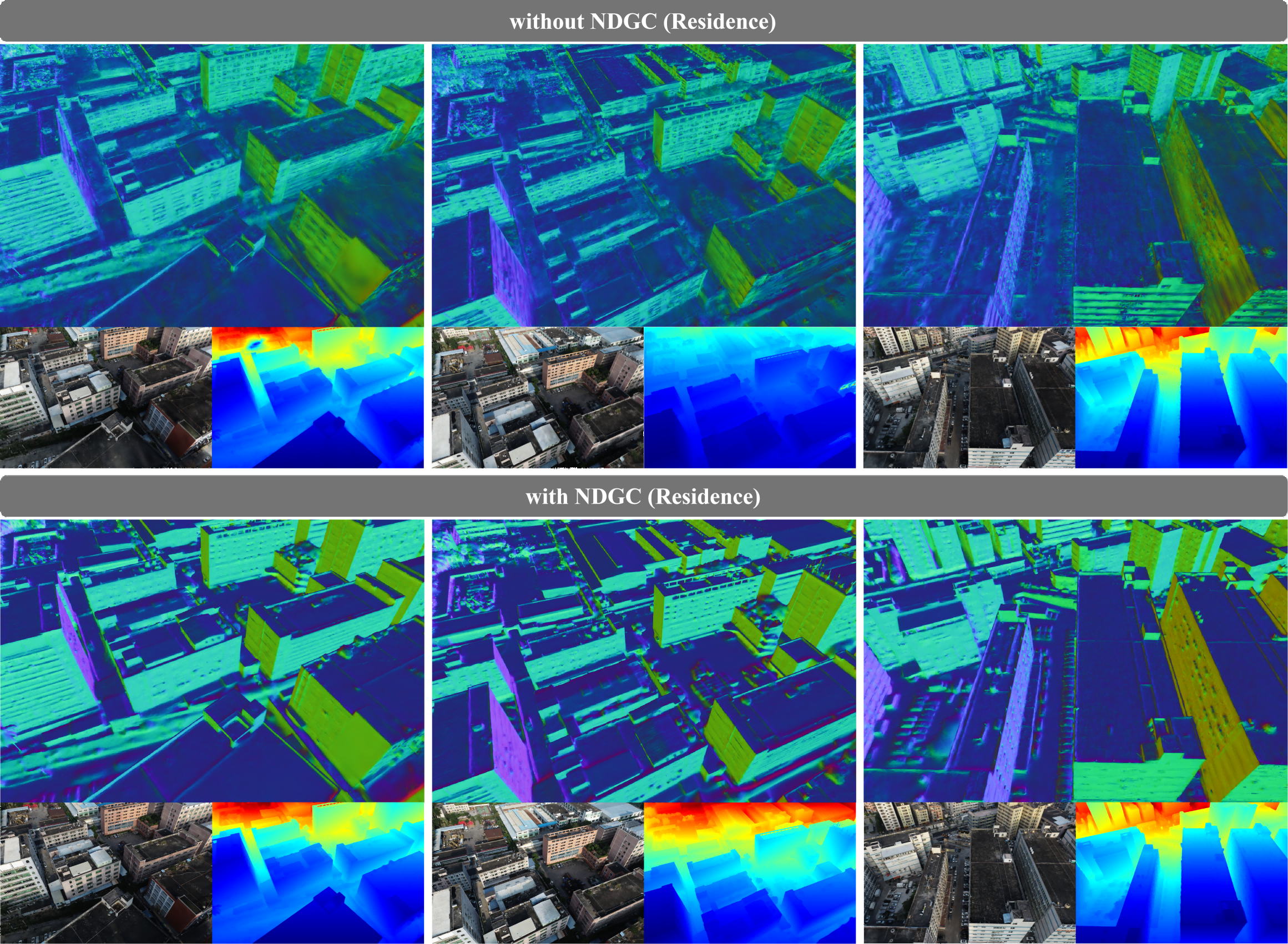}
    \caption{\textbf{DNGC ablation on the Residence scene.}
    \textit{Top:} Without DNGC, normal maps show fragmented orientations on rooftops and walls, with depth artifacts at structural boundaries.
    \textit{Bottom:} With DNGC, normals become coherent on large planar surfaces, and depth transitions are cleaner at building edges. The edge-aware smoothness term preserves sharp boundaries between adjacent structures while enforcing planarity within each surface.}
    \label{fig:app_dngc_residence}
\end{figure*}

We present qualitative ablations of the DNGC regularization on the Building and Residence scenes. Figs.~\ref{fig:app_dngc_building} and~\ref{fig:app_dngc_residence} compare normal maps and depth maps with and without DNGC.

\section{China-Pagoda Dataset Details}
\label{sec:app_pagoda}

\begin{figure*}[p]
    \centering
    \includegraphics[width=\linewidth]{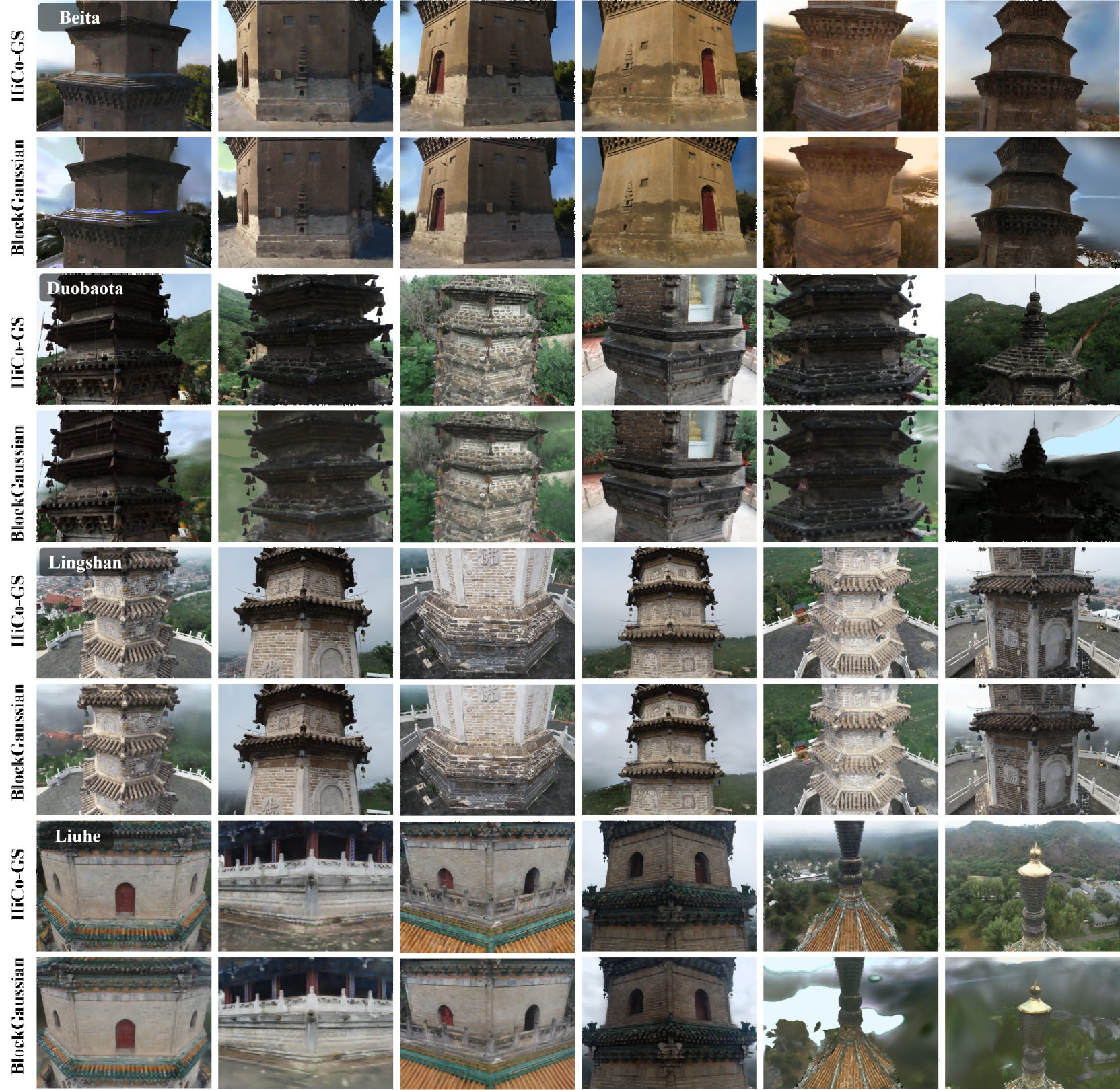}
    \caption{\textbf{China-Pagoda sample images (Part 1):} Beita, Duobaota, Lingshan, and Liuhe pagodas shown from multiple drone viewpoints. These pagodas feature octagonal and hexagonal cross-sections, tiered eaves with hanging bells, intricate brick carvings, and arched doorways.}
    \label{fig:app_pagoda_gallery1}
\end{figure*}

\begin{figure*}[p]
    \centering
    \includegraphics[width=\linewidth]{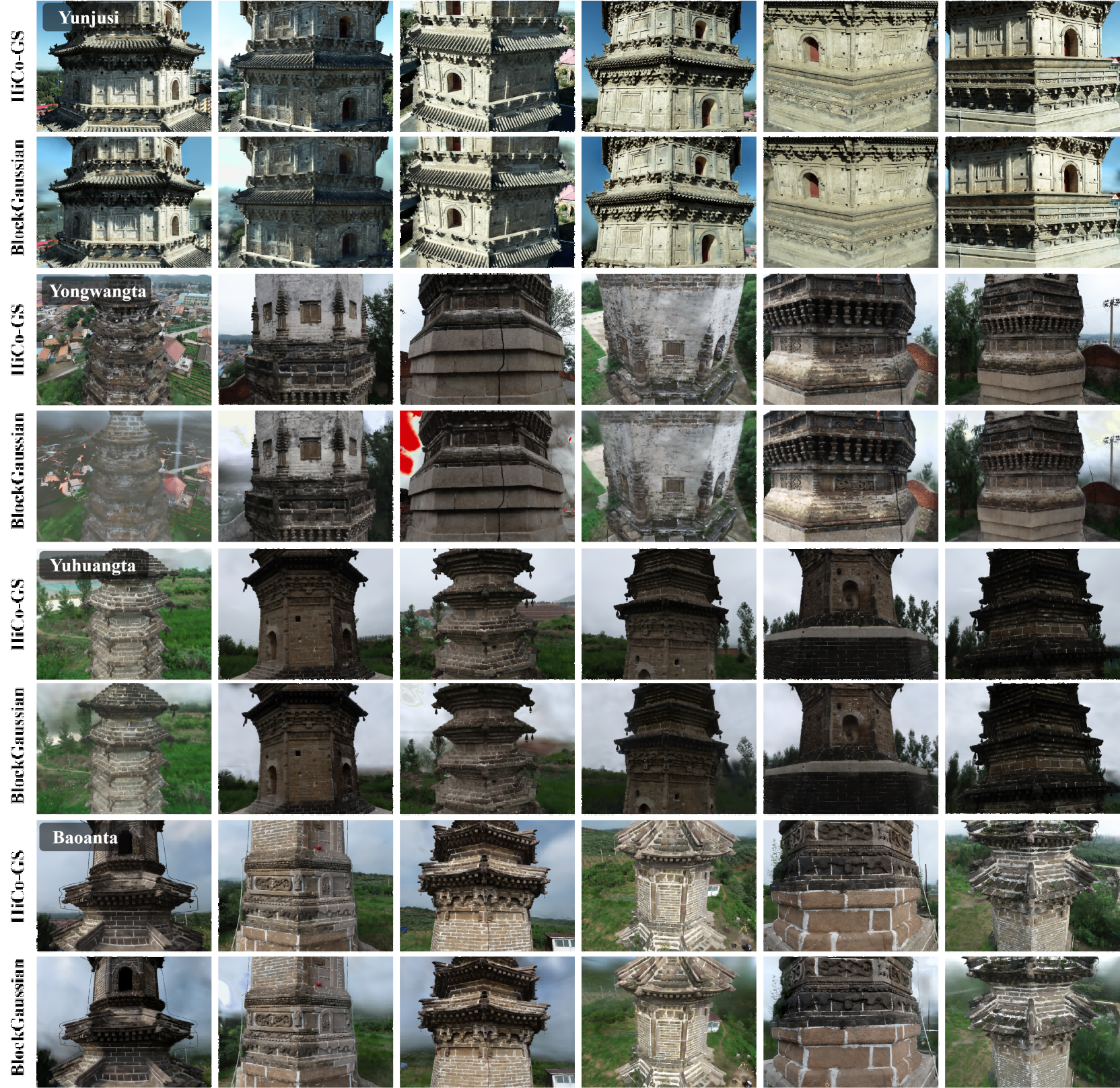}
    \caption{\textbf{China-Pagoda sample images (Part 2):} Yunjusi, Yongwangta, Yuhuangta, and Baoanta pagodas. These structures exhibit heavily weathered surfaces, dense repetitive textures from layered brick construction, and fine-grained ornamental details at varying scales.}
    \label{fig:app_pagoda_gallery2}
\end{figure*}

\begin{figure*}[p]
    \centering
    \includegraphics[width=\linewidth]{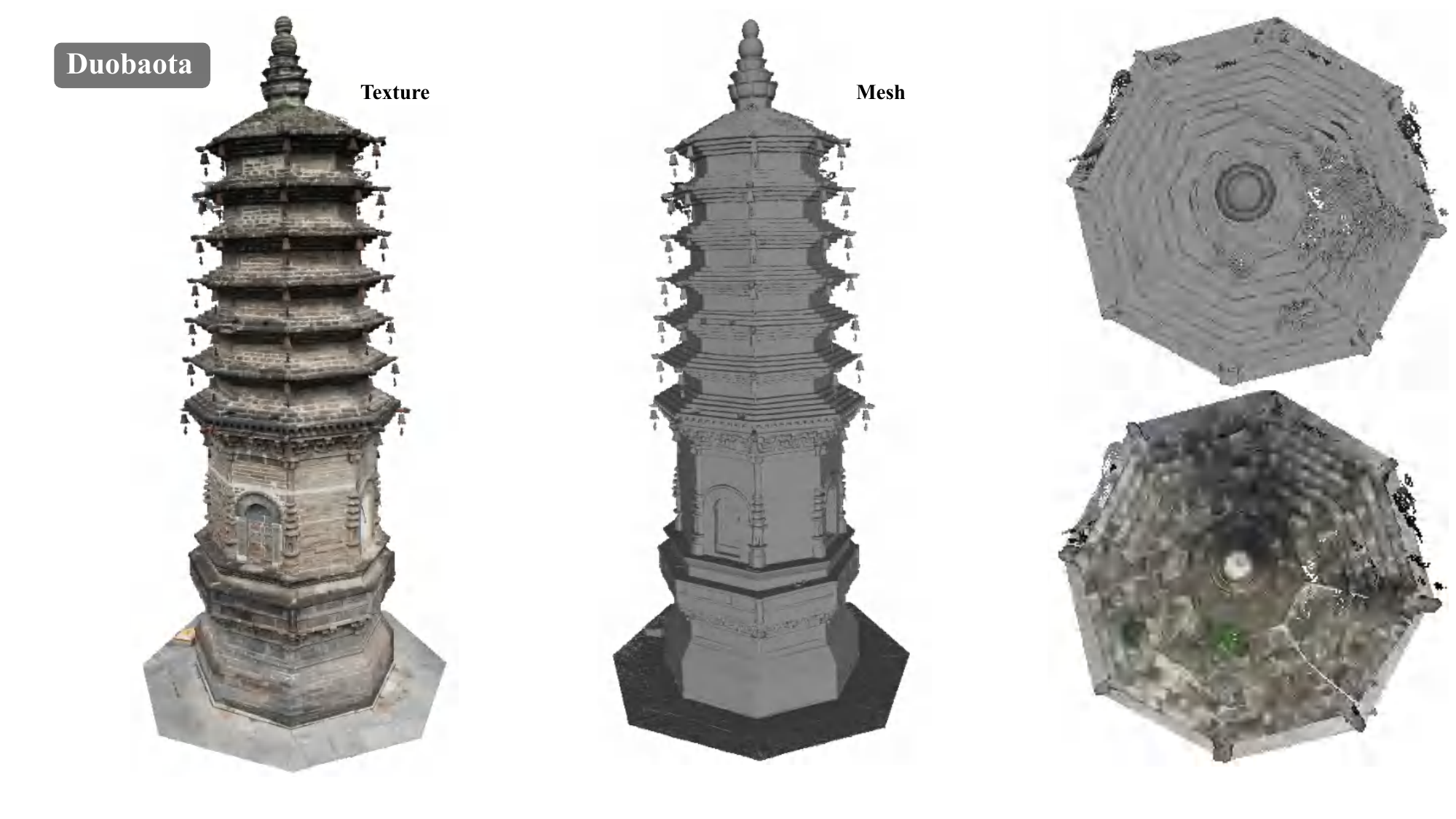}
    \caption{\textbf{Reconstruction of Duobaota.} From left to right: textured rendering, extracted mesh, and top-down views. The multi-layer eaves, hanging bells, and hexagonal base geometry are faithfully reconstructed with clean surface normals and sharp edge boundaries.}
    \label{fig:app_pagoda_duobaota}

    \vspace{4pt}

    \includegraphics[width=\linewidth]{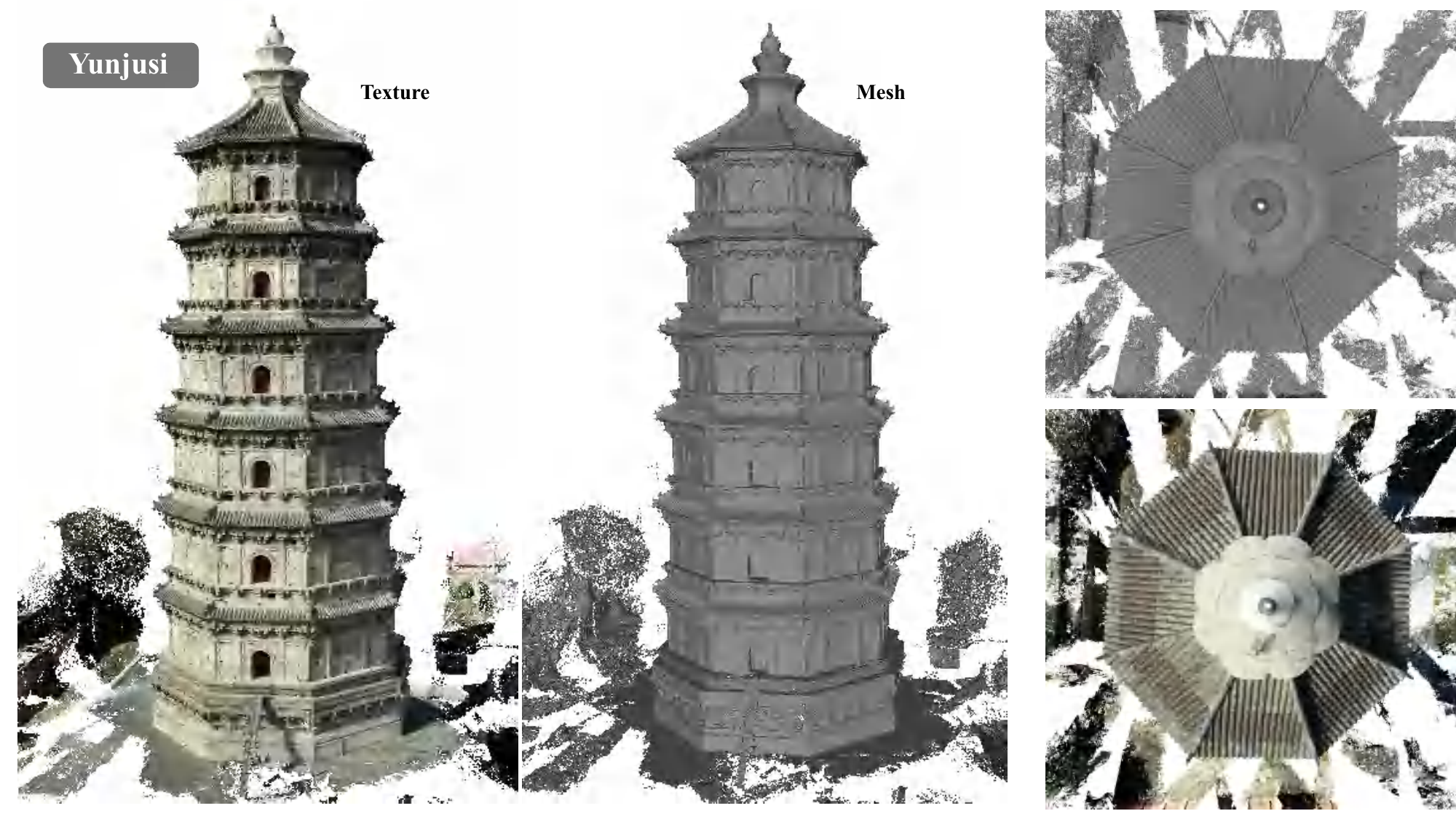}
    \caption{\textbf{Reconstruction of Yunjusi.} The tallest pagoda in our dataset with seven tiers of eaves and dense window openings. The mesh preserves octagonal symmetry and layered eave overhangs, with top-down views revealing clean rooftop geometry and radial ridge lines.}
    \label{fig:app_pagoda_yunjusi}
\end{figure*}

\begin{figure*}[p]
    \centering
    \includegraphics[width=\linewidth]{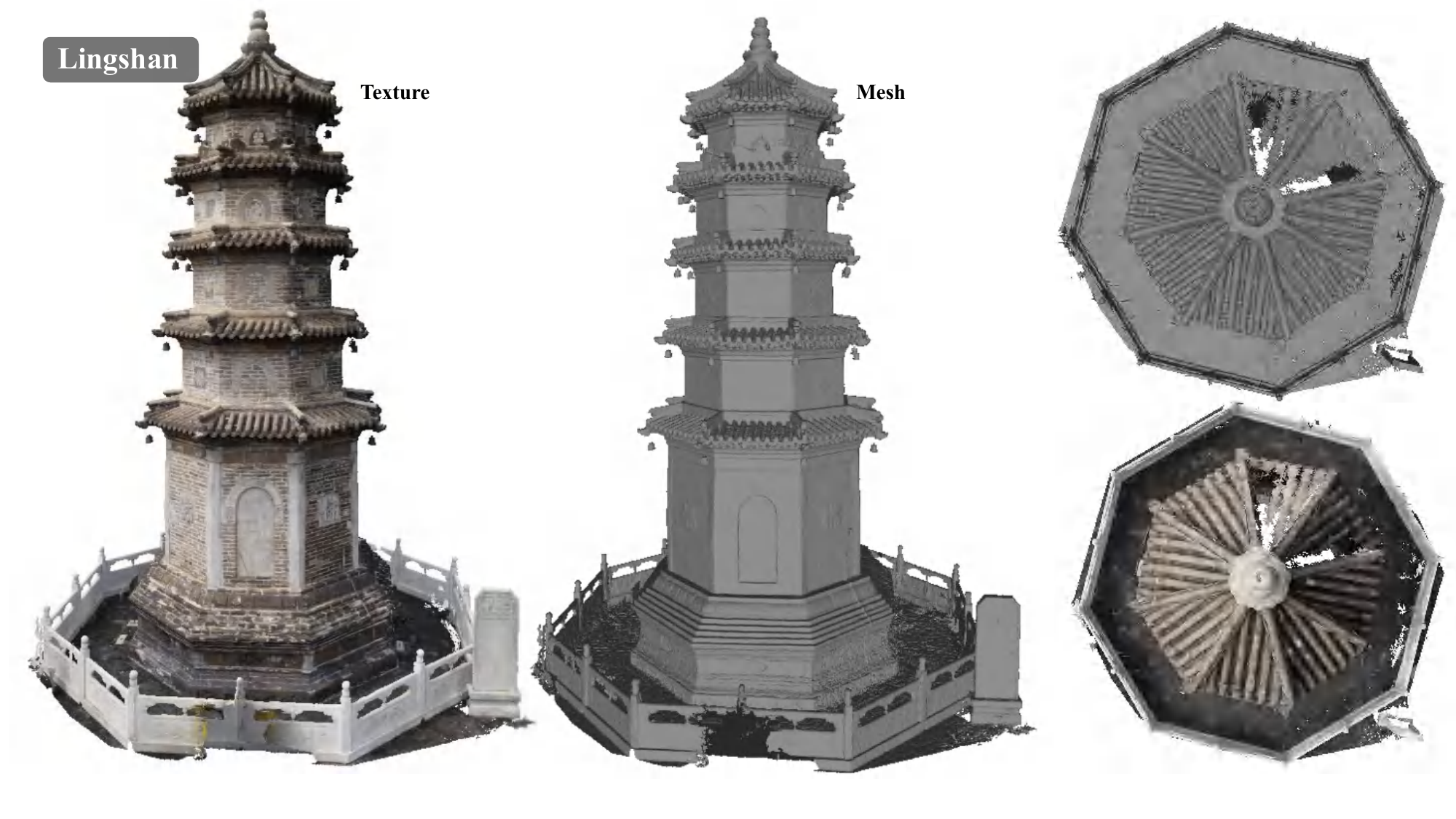}
    \caption{\textbf{Reconstruction of Lingshan.} This pagoda features prominent curved eaves with decorative bells and a surrounding stone balustrade. The mesh accurately captures eave curvature, the octagonal tiered structure, and fine railing geometry at the base.}
    \label{fig:app_pagoda_lingshan}

    \vspace{4pt}

    \includegraphics[width=\linewidth]{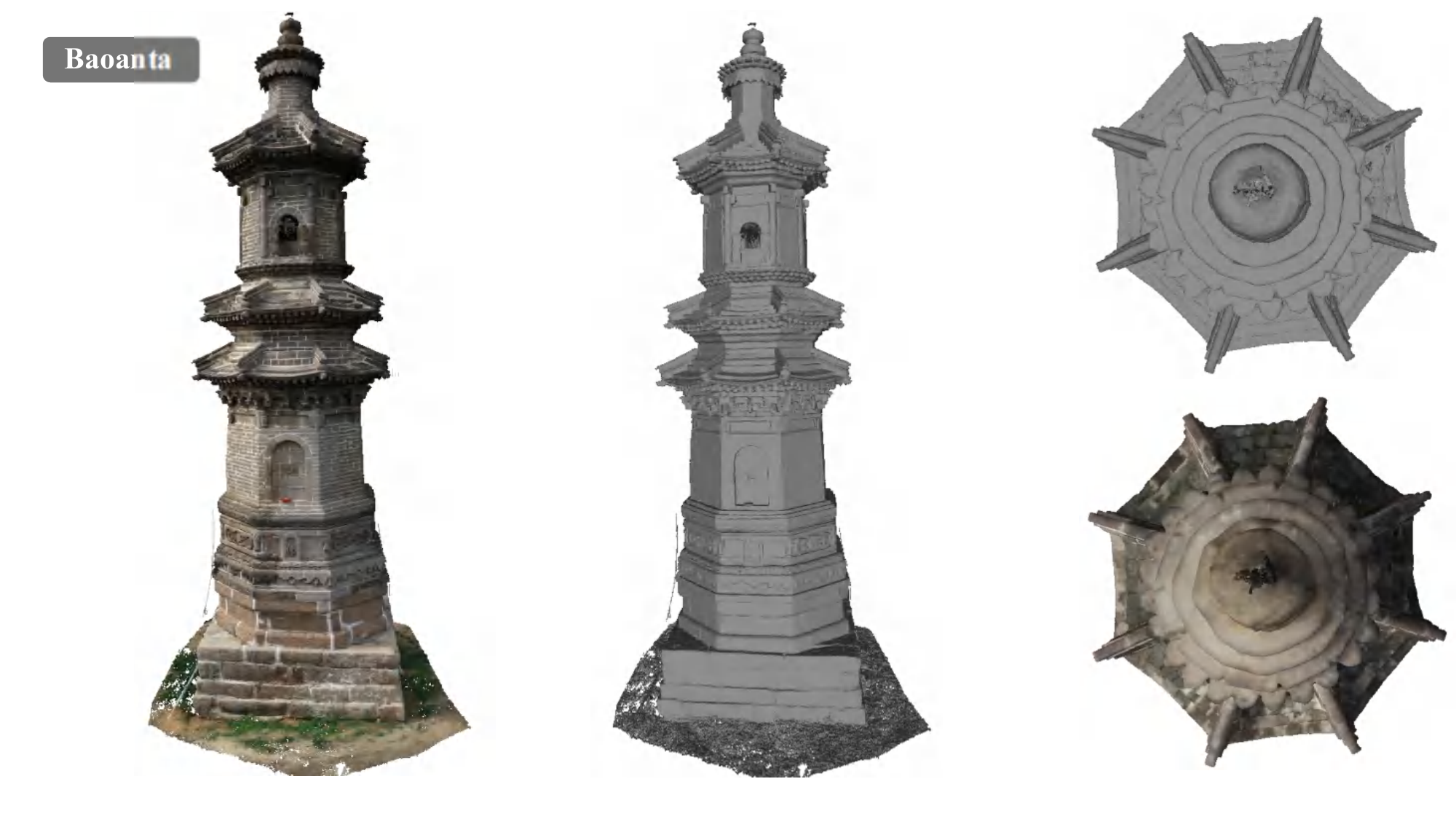}
    \caption{\textbf{Reconstruction of Baoanta.} Textured rendering, extracted mesh, and top-down views demonstrating faithful recovery of the pagoda's layered structure and surface detail.}
    \label{fig:app_pagoda_baoanta}
\end{figure*}

\begin{figure*}[p]
    \centering
    \includegraphics[width=\linewidth]{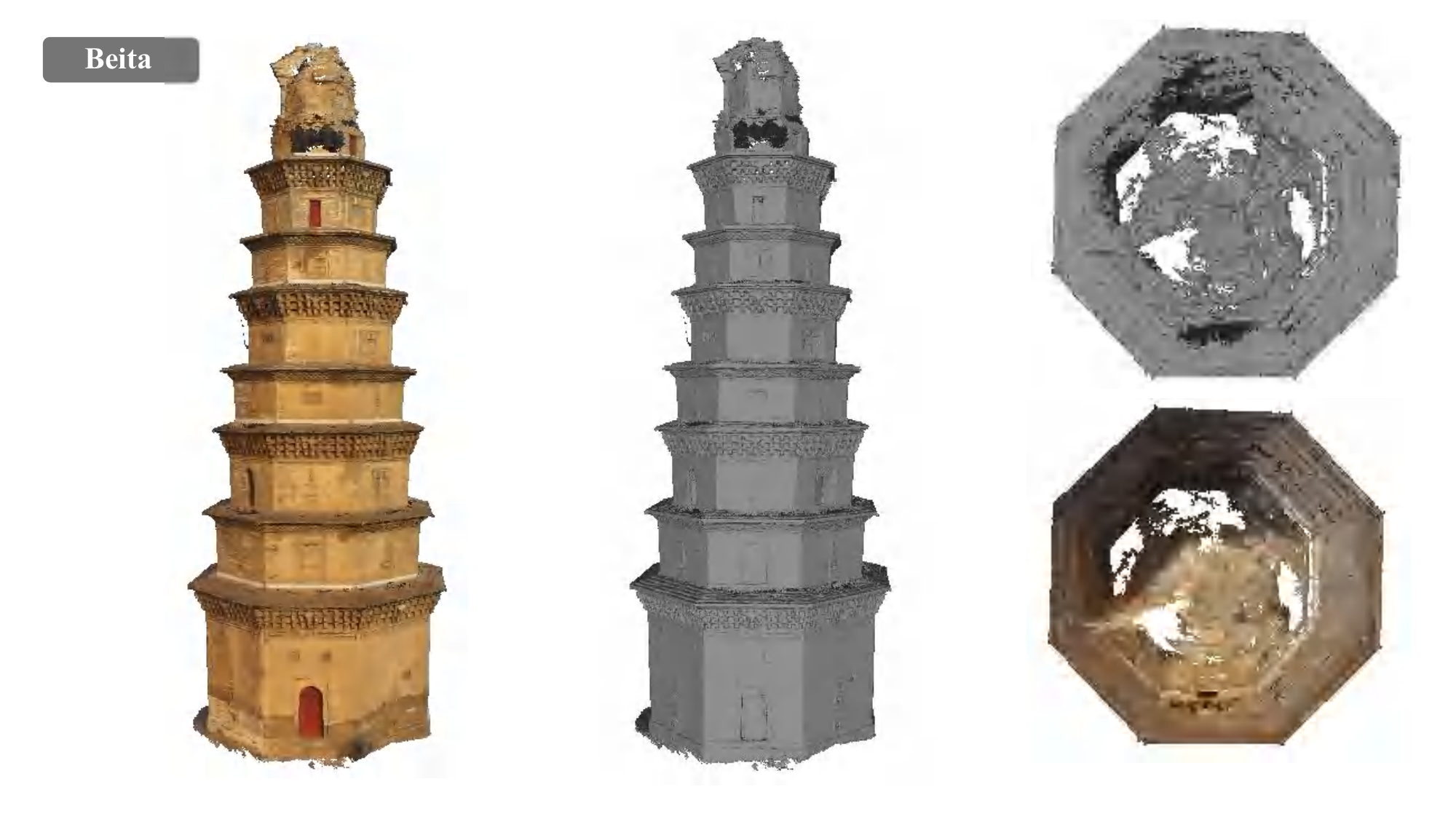}
    \caption{\textbf{Reconstruction of Beita.} The dense brick carvings and weathered surface patterns are well preserved in both the rendered appearance and the extracted mesh geometry.}
    \label{fig:app_pagoda_beita}

    \vspace{4pt}

    \includegraphics[width=\linewidth]{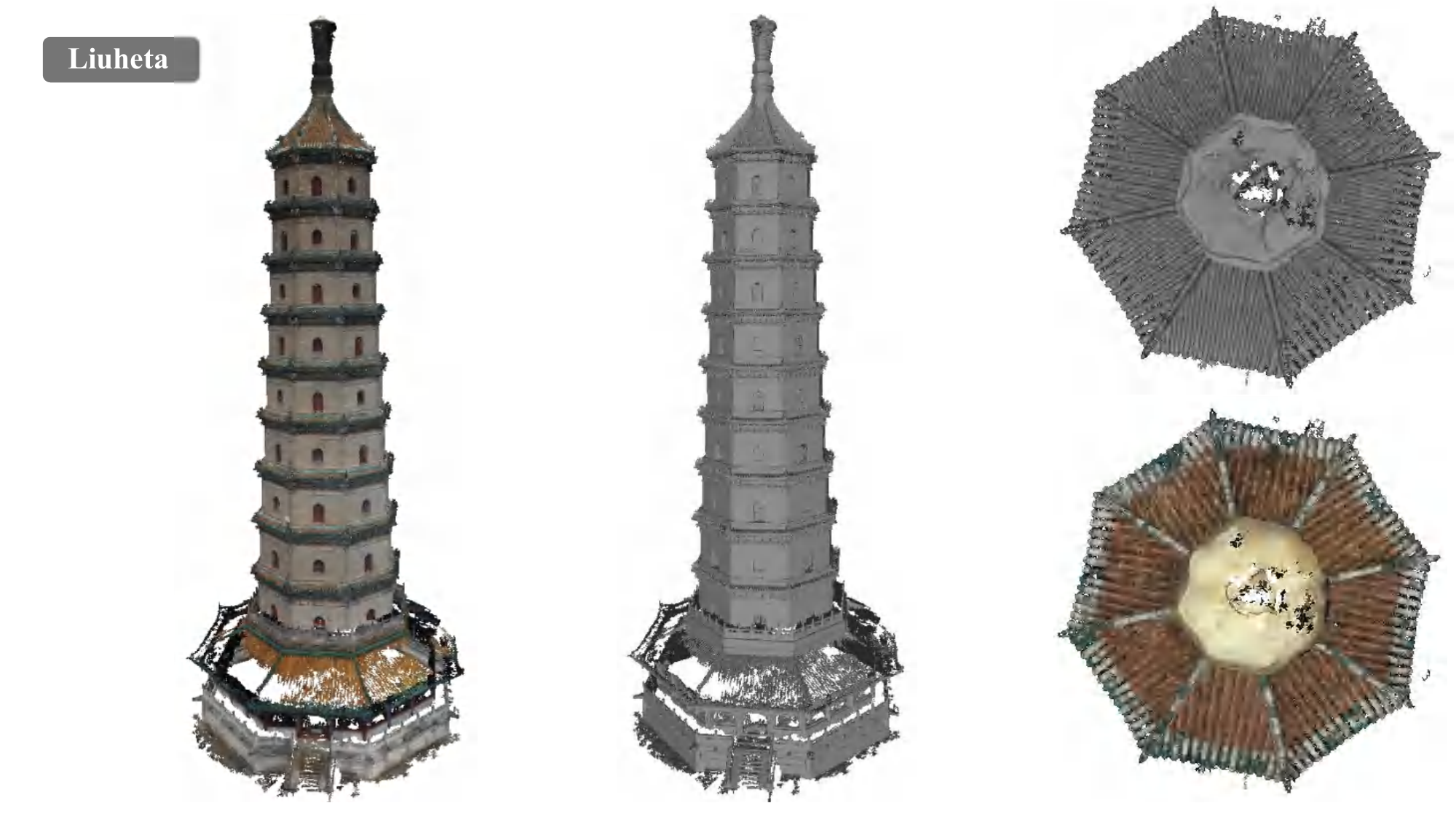}
    \caption{\textbf{Reconstruction of Liuheta.} The multi-story structure with alternating eave tiers and window openings is reconstructed with consistent geometry across all levels.}
    \label{fig:app_pagoda_liuhe}
\end{figure*}

\begin{figure*}[p]
    \centering
    \includegraphics[width=\linewidth]{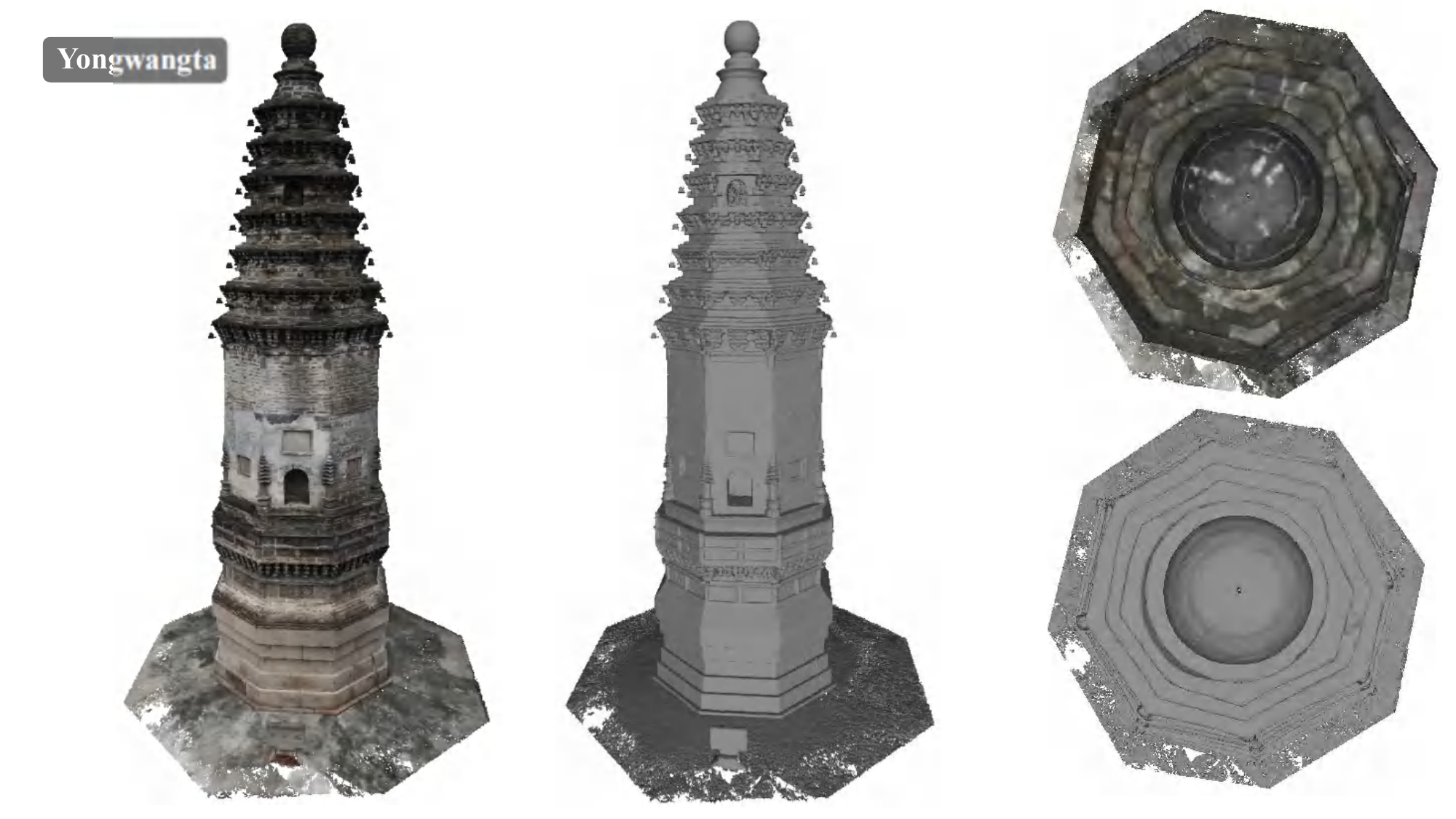}
    \caption{\textbf{Reconstruction of Yongwangta.} Clean mesh extraction captures the tapered silhouette and tiered eave overhangs characteristic of this architectural style.}
    \label{fig:app_pagoda_yongwang}

    \vspace{4pt}

    \includegraphics[width=\linewidth]{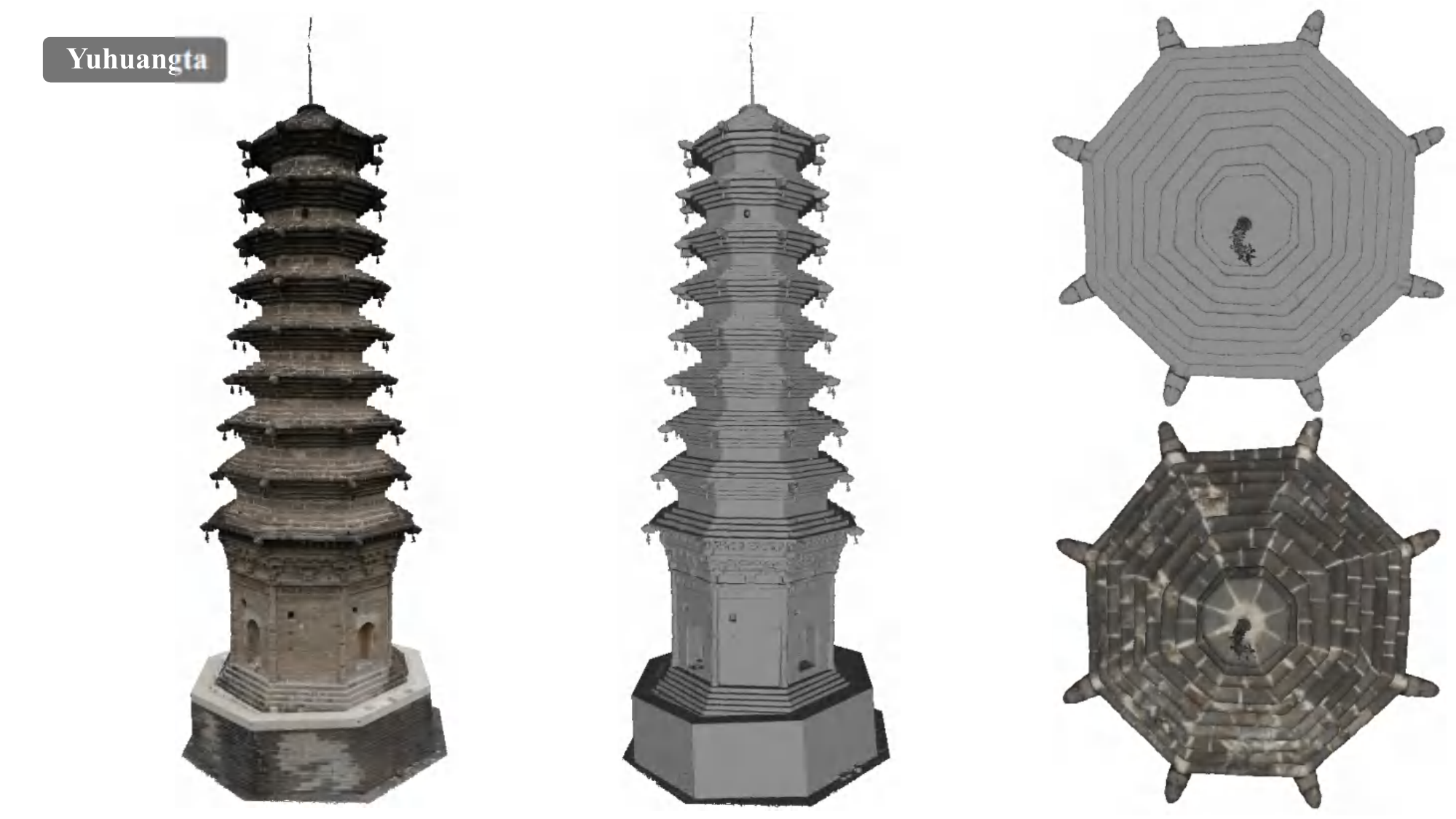}
    \caption{\textbf{Reconstruction of Yuhuangta.} Fine ornamental details and the glazed-tile surface texture are faithfully reproduced across multiple viewpoints.}
    \label{fig:app_pagoda_yuhuang}
\end{figure*}

The China-Pagoda dataset contains 8 ancient Chinese pagodas, each captured with over 1{,}200 drone and ground-level images. Figs.~\ref{fig:app_pagoda_gallery1} and~\ref{fig:app_pagoda_gallery2} present sample images illustrating the diversity of architectural styles and the challenging visual characteristics: dense ornamental carvings, curved multi-layer eaves, repetitive brick textures, and weathered stone surfaces. Figs.~\ref{fig:app_pagoda_duobaota}--\ref{fig:app_pagoda_yuhuang} show detailed reconstruction results including textured renders, extracted meshes, and top-down views.

\section{Large-Scale Urban Scene Comparison}
\label{sec:app_urban}

\begin{figure*}[p]
    \centering
    \includegraphics[width=0.85\linewidth]{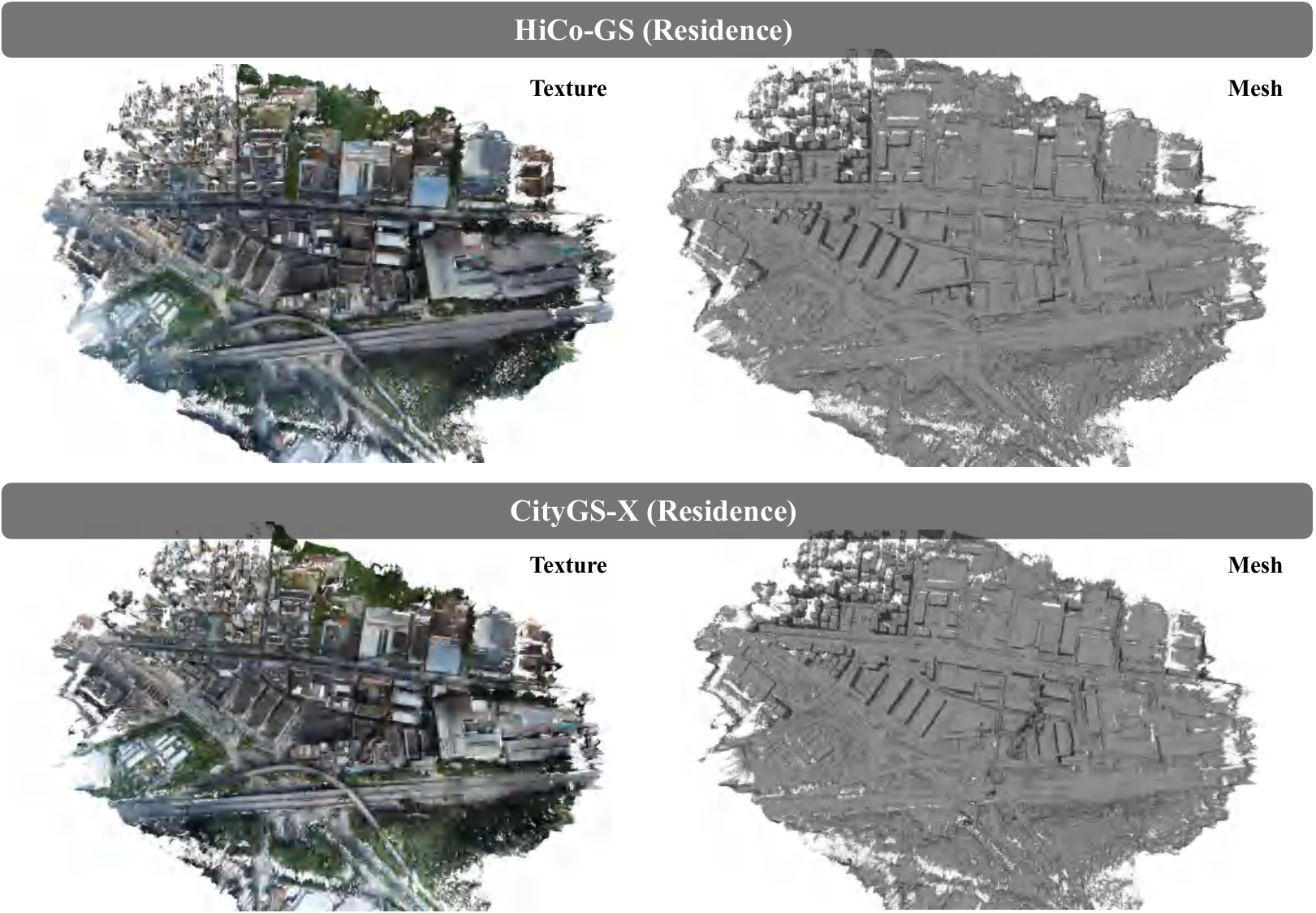}
    \caption{\textbf{Full-scene comparison on Residence.}
    \textit{Top:} HiCo-GS produces consistent facade coloring and clean planar surfaces on rooftops and roads.
    \textit{Bottom:} CityGS-$\mathcal{X}$ exhibits more floating artifacts and noisier mesh geometry, especially around vegetation boundaries and overpass structures.}
    \label{fig:app_urban_residence}

    \vspace{4pt}

    \includegraphics[width=0.85\linewidth]{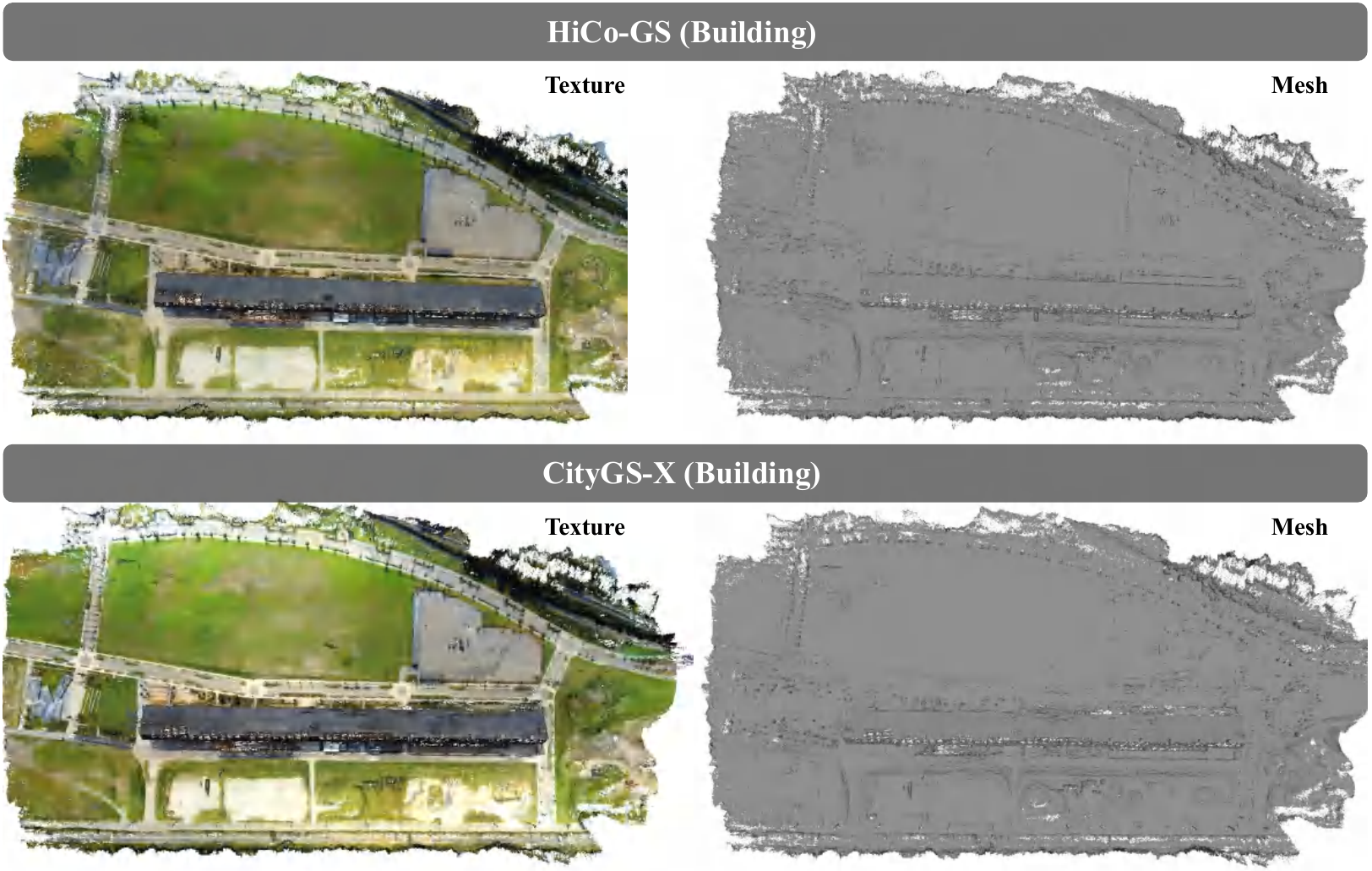}
    \caption{\textbf{Full-scene comparison on Building.}
    \textit{Top:} HiCo-GS achieves well-defined building edges and smooth ground planes.
    \textit{Bottom:} CityGS-$\mathcal{X}$ shows more surface noise on large flat regions such as the sports field and rooftops, where the lack of cross-level context and geometric consistency leads to less coherent surface reconstruction.}
    \label{fig:app_urban_building}
\end{figure*}

Figs.~\ref{fig:app_urban_residence} and~\ref{fig:app_urban_building} compare HiCo-GS against CityGS-$\mathcal{X}$ on the UrbanScene3D Residence and Building scenes with full-scene textured renderings and extracted meshes.

\end{document}